\documentclass[10pt,twocolumn,letterpaper]{article}
\usepackage[pagenumbers]{cvpr}

\usepackage{graphicx}
\usepackage{booktabs}
\usepackage{xcolor}
\usepackage{colortbl}
\usepackage{multirow}
\usepackage{algorithmic}
\usepackage{algorithm}
\usepackage{amsmath}
\usepackage{pgfplots}
\usepackage{tabularx}
\usepackage{wrapfig}
\usepackage{placeins}
\pgfplotsset{width=3cm,compat=1.9}

\usepackage[accsupp]{axessibility}  

\definecolor{tabthird}{HTML}{ffffcc}
\definecolor{tabsecond}{HTML}{d9f0a3}
\definecolor{tabfirst}{HTML}{78c679}

\definecolor{colorTabTop}{rgb}{0.95,0.93,0.9}
\definecolor{colorTab}{rgb}{0.9,0.97,0.9}
\definecolor{color3}{rgb}{0.95,0.95,0.95}

\usepackage{hyperref}

\usepackage{orcidlink}

\title{\vspace{-8pt}Telescope:\\ Learnable Hyperbolic Foveation for Ultra-Long-Range Object Detection}

\author{Parker Ewen$^{1}$, Dmitriy Rivkin$^{1}$, Mario Bijelic$^{1,2}$, Felix Heide$^{1,2}$ \\[6pt]
$^1$Torc Robotics, $^2$Princeton University \\ {\normalsize }
}

\begin{document}
 \maketitle

\begin{abstract}
Autonomous highway driving, especially for long-haul heavy trucks, requires detecting objects at long ranges beyond 500 meters to satisfy braking distance requirements at high speeds. At long distances, vehicles and other critical objects occupy only a few pixels in high-resolution images, causing state-of-the-art object detectors to fail. 
This challenge is compounded by the limited effective range of commercially available LiDAR sensors, which fall short of ultra-long range thresholds because of quadratic loss of resolution with distance, making image-based detection the most practically scalable solution given commercially available sensor constraints.
We introduce Telescope, a two-stage detection model designed for ultra-long range autonomous driving. 
Alongside a powerful detection backbone, this model contains a novel re-sampling layer and image transformation to address the fundamental challenges of detecting small, distant objects. 
Telescope achieves $76\%$ relative improvement in mAP in ultra-long range detection compared to state-of-the-art methods (improving from an absolute mAP of 0.185 to 0.326 at distances beyond 250 meters), requires minimal computational overhead, and maintains strong performance across all detection ranges. Our project page is available at \url{https://light.princeton.edu/telescope}.
\end{abstract} \label{sec:abs}
\section{Introduction}

Autonomous driving requires perceptual understanding of the surrounding scene \cite{yurtsever2020survey, wong2020mapping}. Existing datasets and benchmarks focus heavily on city driving \cite{geiger2012kitti, caesar2020nuscenes, chang2019argoverse, sun2020waymo, huang2018apolloscape, wilson2023argoverse, cordts2016cityscapes}, where low vehicle speeds require short-range perception.

Highway driving, and in particular heavy-duty trucking, presents a fundamentally different challenge. At highway speeds, a fully-loaded truck requires on the order of $150$--$200\,\mathrm{m}$ to come to a complete stop~\cite{ghilotti2026truckdrive}.
Existing benchmarks often provide limited sensing horizons around $80$--$100\,\mathrm{m}$, which is insufficient for safe braking and strategic maneuvers such as merging or lane changes.
As a result, safe highway autonomy requires reliable perception at hundreds of meters, and up to the kilometer scale for full visual scene understanding \cite{ghilotti2026truckdrive}. 
\begin{figure}[t]
    \centering
    \includegraphics[width=\linewidth]{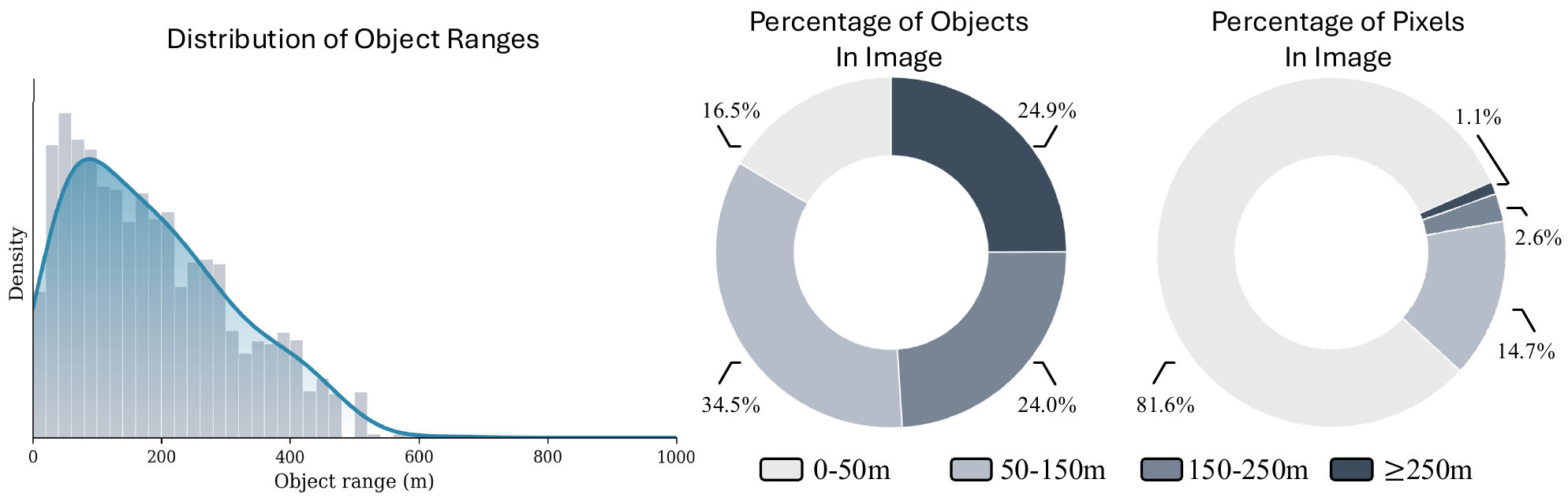}
    \caption{\textbf{Long-range Objects in Driving Datasets.} Analysis of the TruckDrive~\cite{ghilotti2026truckdrive} dataset shows the distribution of object distances and the breakdown of the pixel-wise composition of objects at each distance. While all object ranges are equally represented in images, the proportion of pixel area disproportionately favors nearby objects, with long ($150-250$m) and ultra-long ($\geq 250$m) objects occupying only a small fraction of image pixels.}
    \label{fig:dataset_analysis}
\end{figure}
We study ultra-long range object detection in TruckDrive \cite{ghilotti2026truckdrive} which provides annotations up to $1\,\mathrm{km}$.
From this dataset, we identify three major challenges for ultra-long range detection. First, active LiDAR and radar measurements become increasingly sparse and low in signal at long distances ---fundamental to the sequential scanning and quadratic intensity falloff of diffuse reflections---making cameras the only sensing modality that can provide dense spatial coverage at hundreds of meters.
Second, ultra-long range objects project to extremely small bounding boxes. 
In these datasets, objects beyond $250\mathrm{m}$ frequently occupy only tens of pixels.
Consequently, the number of resolvable visual features on distant objects is directly limited by the input image resolution.
Third, there is an extreme scale imbalance between objects, where nearby objects and background regions dominate the pixel budget and ultra-long range objects contribute only a negligible fraction of tokens or patches (Figure \ref{fig:dataset_analysis}).

Figure~\ref{fig:dataset_analysis} illustrates this imbalance, where images in TruckDrive contain both nearby and distant objects.
Notably, distant objects represent a fraction of the image features compared to nearby objects.
This distribution imbalance highlights that ultra-long range detection is not only a long-tail data problem, but also a fundamental representational problem caused by extreme scale disparity within individual images.

Detection methods must therefore efficiently process high-resolution images with minimal latency and memory usage in order to capture objects at long range. 
We argue the detection mechanism should therefore avoid the quadratic complexity of standard self-attention, which becomes prohibitive at high resolutions and dilutes attention \cite{carion2020detr, zhu2020deformable}.
Finally, the model requires an explicit resampling mechanism that magnifies distant objects while shrinking nearby ones, normalizing object scale to facilitate learning \cite{jaderberg2015spatial}.

\begin{figure*}[h]
    \centering
    \includegraphics[width=0.95\textwidth]{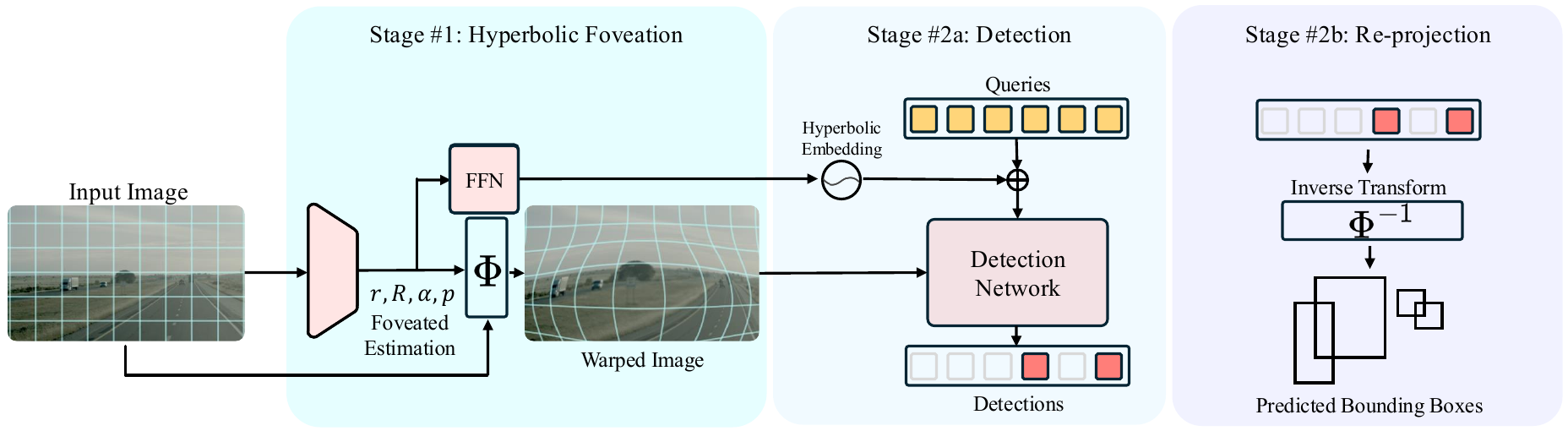}
    \caption{\textbf{Telescope.} We propose a two-stage ultra-long range detection model. Stage one uses a down-sampled image to estimate the hyperbolic foveation image transformation parameters. This transformation enlarges distant objects at the center of the transform while shrinking nearby objects at the periphery. Stage two uses this transformed image alongside learned hyperbolic embeddings to detect objects at distances of up to 1km.}
    \label{fig:network_diagram}
\end{figure*}

To address these requirements, we introduce Telescope, a two-stage ultra-long range detection method with an application to autonomous highway driving (Figure \ref{fig:network_diagram}). 
In the first stage, we learn an image transformation called the hyperbolic foveation, which magnifies salient image regions. 
This transformation interpolates between the Poincaré disk projection \cite{stanoyevitch1994geometry} and the identity function, inducing a Riemannian space \cite{lee2018introduction} where bounding boxes can be parameterized and learned, then re-projected to image space with machine precision and no warping artifacts.
Notably, this transform is inspired by biological vision~\cite{jabbireddy2022foveated}.
By learning the foveation parameters on down-sampled images, stage one incurs minimal computational overhead.

The second stage applies the learned transformation to full-resolution images, which are then processed by a pre-trained foundation model encoder \cite{ravi2024sam, carion2025sam, oquab2023dinov2, simeoni2025dinov3, bolya2025perception} and lightweight Deformable DETR detection head \cite{zhu2020deformable}. 
The pre-trained encoder enables fast convergence and robustness across scales, while sparse sampling avoids the quadratic cost and attention dilution of standard transformers, making it well-suited for high-resolution inputs.
While this two-stage approach is applied to the context of ultra-long range object detection for highway driving, the proposed foveated transform is general and can be applied to existing image-based approaches, including Vision Language Models (VLMs).

Evaluated on long-range autonomous driving benchmarks, Telescope consistently improves detection performance across all distance ranges and achieves up to $76\%$ improvement in mAP at ultra-long ranges  over existing state-of-the-art approaches (improving from an absolute mAP of 0.185 to 0.326 at distances greater than $250\mathrm{m}$), with detections extending to $1\,\mathrm{km}$.
In summary, the contributions of this paper are:
\begin{itemize}
    \item An analysis of ultra-long range object detection and the identification of several critical model requirements.
    \item A systematic ablation of foundation model image encoders, detection heads, and training schemes for ultra-long range object detection, providing practical insights into which backbone representations and optimization strategies are most effective under constrained fine-tuning budgets.
    \item A novel learnable and invertible hyperbolic foveation image transform and associated Riemannian bounding box reparameterization for ultra-long range domain scaling.
    \item A state-of-the-art ultra-long range object detection model, Telescope, for highway driving, which improves object detection performance by $76\%$ at ultra-long ranges compared to existing methods (increasing absolute mAP from 0.185 to 0.326 at distances beyond 250 meters). 
\end{itemize} \label{sec:intro}
\section{Related Work}
Object detection has been a cornerstone application of modern deep learning-based computer vision over the past decade \cite{zou2023object, zhao2019object}. 
The autonomous driving domain, in particular, has spurred innovations in detection methods tailored to the unique challenges of on-road perception \cite{mao20233d, ren2015faster}. 
We present a review of relevant work for object detection for autonomous driving, small object detection, and learned spatial transformations.

\paragraph{Object Detection for Autonomous Driving.}
Autonomous driving has helped push the development of specialized object detection benchmarks and methods. 
Prominent datasets include KITTI~\cite{geiger2012kitti}, nuScenes~\cite{caesar2020nuscenes}, Waymo Open Dataset~\cite{sun2020waymo}, CityScapes~\cite{cordts2016cityscapes}, and Argoverse~\cite{chang2019argoverse, wilson2023argoverse}, which provide multi-modal sensor data primarily catered towards low-speed and city driving scenarios.
Early methods adapted general-purpose detectors like Faster R-CNN~\cite{ren2015faster, chen2016r} and YOLO~\cite{redmon2016yolo} to automotive contexts. 
More recent approaches leverage transformer-based architectures, including DETR~\cite{carion2020detr} and its variants \cite{dai2021dynamic, li2022dn, dai2021up}, which formulate detection as a set prediction problem. 
Deformable DETR~\cite{zhu2020deformable} introduced sparse spatial sampling \cite{xia2022vision} to improve efficiency and convergence for high-resolution inputs. 
Methods like CenterNet~\cite{zhou2019objects} and FCOS~\cite{tian2019fcos} explore anchor-free detection paradigms better suited to the wide range of object scales in driving scenes. 
Despite these advances, most benchmarks emphasize urban driving scenarios at limited ranges \cite{cordts2016cityscapes, sun2020waymo}, leaving ultra-long range highway detection under-explored.

\paragraph{Small Object Detection.}
Small object detection presents fundamental challenges stemming from limited pixel support \cite{nguyen2020evaluation}, poor signal-to-noise ratios \cite{lee2021self}, and severe class imbalance in high-resolution images \cite{wei2024review, mirzaei2023small}.
Standard detection architectures struggle when objects occupy few pixels due to metrics such as IoU scaling poorly to small bounding boxes~\cite{cheng2023survey, guo2023save}. 
Methods addressing small objects typically employ multi-scale feature pyramids~\cite{lin2017feature, gong2021effective, yang2022querydet}, specialty losses \cite{xu2022rfla, wang2021normalized}, super-resolution preprocessing~\cite{bai2018finding, noh2019better, liu2024small}, or attention mechanisms to enhance fine-grained representations~\cite{chen2020mmdetection, guo2023save, wang2022advancing, chen2016r}. 

Datasets like TinyPerson~\cite{yu2020scale}, AI-TOD~\cite{xu2022detecting}, and DOTA~\cite{xia2018dota} focus specifically on small object scenarios, primarily in aerial imagery and crowd surveillance contexts.
However, these datasets either contain only small objects \cite{xu2022detecting, xia2018dota} or tend to have limited variance in object sizes within individual images \cite{yu2020scale}.
In contrast, object sizes vary significantly in autonomous driving, where nearby vehicles may occupy orders-of-magnitude more pixels than distant vehicles \cite{ghilotti2026truckdrive}. 

\paragraph{Learned Spatial Transformations.}
While some approaches focus on network architectures and loss functions tailored towards the small object domain \cite{wang2021normalized, de2023unbalanced, zhang2022dino, guo2023save, gong2021effective}, other methods re-sample the image directly to magnify or warp high salience regions.
Spatial Transformer Networks~\cite{jaderberg2015spatial} introduced learnable geometric transformations to warp input images for improved spatial invariance. 
More recently, FOVEA~\cite{thavamani2021fovea} extends this concept to autonomous driving with a learned foveation that magnifies distant regions for long-range detection. 
However, FOVEA requires full-resolution images to estimate transformation parameters and maintains axis-aligned bounding box representations. 
Inspired by biological visual systems \cite{jabbireddy2022foveated}, our proposed hyperbolic foveation differs by estimating parameters from low-resolution images by leveraging strong object detection priors from the network encoder.
Furthermore, the proposed method directly estimates the warped bounding boxes in the local Riemannian coordinate frame without requiring a rectilinear coordinate frame.
This reparameterization enables more natural object representations in the transformed domain without the geometric constraints of axis-aligned boxes under non-linear transformations. \label{sec:rel_work}
\section{Ultra-Long-Range Detection}

In this section, we describe the proposed ultra-long range detection method, Telescope, as shown in Figure \ref{fig:network_diagram}. 
We first introduce the hyperbolic foveated transform in Sec.~\ref{sec:transform} as a means of normalizing object sizes across scales.
This transform magnifies distant objects while compressing nearby ones and ensures minimal computational overhead. In Sec.~\ref{sec:parameterization}, we describe the parameterization of the bounding boxes in this transformed space. Finally, we describe a network architecture in Sec.~\ref{sec:network} to efficiently operate on these high-resolution transformed images, enabling object detection at distances of up to 1km.

\subsection{Hyperbolic Foveated Transform}\label{sec:transform}

Let $\mathcal{M}=\{x\in\mathbb{R}^2:\|x\|<1\}$ denote the Poincar\'e disk equipped with the standard metric tensor $ds^2$.
Directly projecting an image with finite domain onto $\mathcal{M}$ is undesirable since the metric diverges as $\|x\|\!\rightarrow\!1$, causing unbounded distortion and numerical instability near the image boundary.

To overcome this issue, we define a pseudo-Riemannian projection which radially interpolates between a Poincar\'e-like contraction and the identity transform.
Let normalized image coordinates be $x\in[-1,1]^2$ and let $o\in[-1,1]^2$ be the Poincar\'e origin projected into Euclidean coordinates.
Define $r=\|x-o\|$ as the offset between the center of the image and the origin of the Poincar\'e disk.
The Poincar\'e projection is given as

\begin{equation}
h(x;o)=o+\frac{\tanh(\alpha r)}{r}(x-o),
\end{equation}

\noindent where $\alpha>0$ is the hyperbolic contraction strength.

The hyperbolic foveated transform is then defined as

\begin{equation}
\label{eq:foveated_map}
\Phi(x)= (1-w(r))\,x + w(r)\,h(x), \qquad
\end{equation}

\noindent where $w(r)=(1-\min(r/R,1))^{p}$ is the radial interpolation coefficient, $p>0$ is the fixed blending exponent, and $R>0$ is the radial scale of the Poincar\'e disk. 
For $r\ll R$, $\Phi$ behaves as a hyperbolic contraction around $o$, while for $r\geq R$ it smoothly approaches the identity mapping.
This enables a numerically stable projection of images onto the induced Riemannian manifold without unbounded distortion, meaning objects near the boundaries of the image are still visible.

While the exact inverse of this transform, $\Phi^{-1}(x)$, cannot be computed explicitly, its existence is provable\footnote{\label{footnote:see_appendix}See Appendix for Theorem and Proof.}.
Furthermore, we can approximate the inverse differentiably, up to numerical precision, and with convergence guarantees\textsuperscript{\ref{footnote:see_appendix}} via the Newton-Raphson algorithm.
Given $y=\Phi(x)$, the inverse $\Phi^{-1}(y)$ is obtained numerically via

\begin{equation}
x^{(k+1)} = x^{(k)} + \eta \big(y-\Phi(x^{(k)})\big),
\label{eq:inv_iter}
\end{equation}

\noindent initialized with $x^{(0)}=y$ and step size $\eta\in(0,1]$.

\subsection{Hyperbolic Box Parameterization}\label{sec:parameterization}
\begin{figure}
    \centering
    \includegraphics[width=0.8\linewidth]{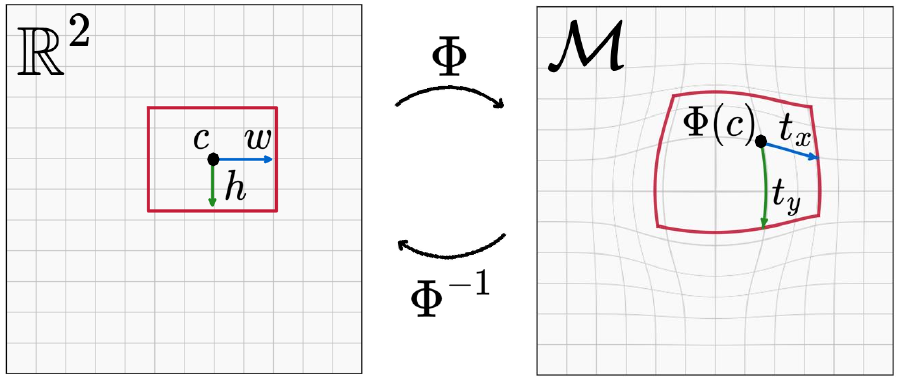}
    \captionof{figure}{\textbf{Hyperbolic Foveated Transform.}
    The transformation coefficients enable the re-parameterization of the bounding box in the induced Riemannian space where the box center and tangent vector magnitudes fully describe the box location and shape.}
    \label{fig:transform}
\end{figure}
When projecting images onto the induced Riemannian manifold, bounding boxes become warped, and axis-aligned, recti-linear boxes can no longer be used.
Akin to the 4-parameter bounding box parameterization in axis-aligned image-space coordinates, 
We propose estimating the local coordinates of these warped boxes directly and provide a 4-parameter parameterization for boxes in the Riemannian manifold induced by Eq.~\eqref{eq:foveated_map}.

Let a box be defined in Euclidean image coordinates by center $c\in\mathbb{R}^2$, width $w\in \mathbb{R}$, and height $h\in \mathbb{R}$ such that $b = [c_x, c_y, w, h].$
The re-parameterized box in the induced Riemannian space is then $b'=[\Phi_x(c), \Phi_y(c), \|t_x\|, \|t_y\|]$, where the first two components are the projected box center and the last two components are the tangent vector magnitudes of the local coordinates at $\Phi(c)$.
Notably, only the tangent vector magnitudes are needed as the tangent vectors are fully defined given the transform parameters.
Figures \ref{fig:transform} and \ref{fig:foveated_examples} provides a visualization of this re-parameterization.

The tangent vectors of the bounding box are computed as
\begin{align}
t_x &= J_\Phi(c)\,[w,0]^\top  \label{eq:tangent_x} \\
t_y &= J_\Phi(c)\,[0,h]^\top  \label{eq:tangent_y}
\end{align}

\noindent where $J_\Phi(c)$ is the Jacobian of $\Phi$ at $c$.

All the terms in the re-parameterization depend only on the original box parameterization and the transform parameters $(\alpha,R,p,o)$, and can therefore be evaluated analytically and vectorized efficiently during training. 
This parameterization is fully determined by $(R,\alpha,p,o)$ through $J_\Phi$ and uniquely specifies the local box geometry in the induced Riemannian space.
Furthermore, the original bounding box parameterization can be recovered using \eqref{eq:inv_iter} and by inverting \eqref{eq:tangent_x} and \eqref{eq:tangent_y}.

\begin{figure}
    \centering
    \includegraphics[width=0.8\linewidth]{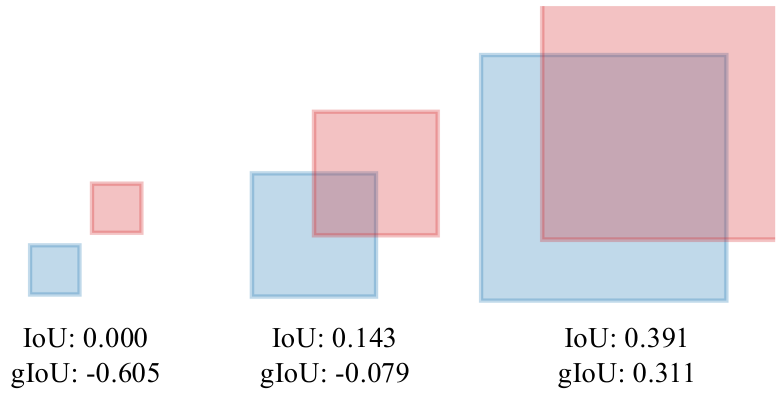}
    \captionof{figure}{\textbf{gIoU for Multi-Scale Training.}
    The gIoU metric is a better training loss for multi-scale objects as it provides a gradient even when bounding boxes are non-overlapping.}
    \label{fig:iou_vs_giou}
\end{figure}
\begin{figure} [t]
    \centering
    \includegraphics[width=\linewidth]{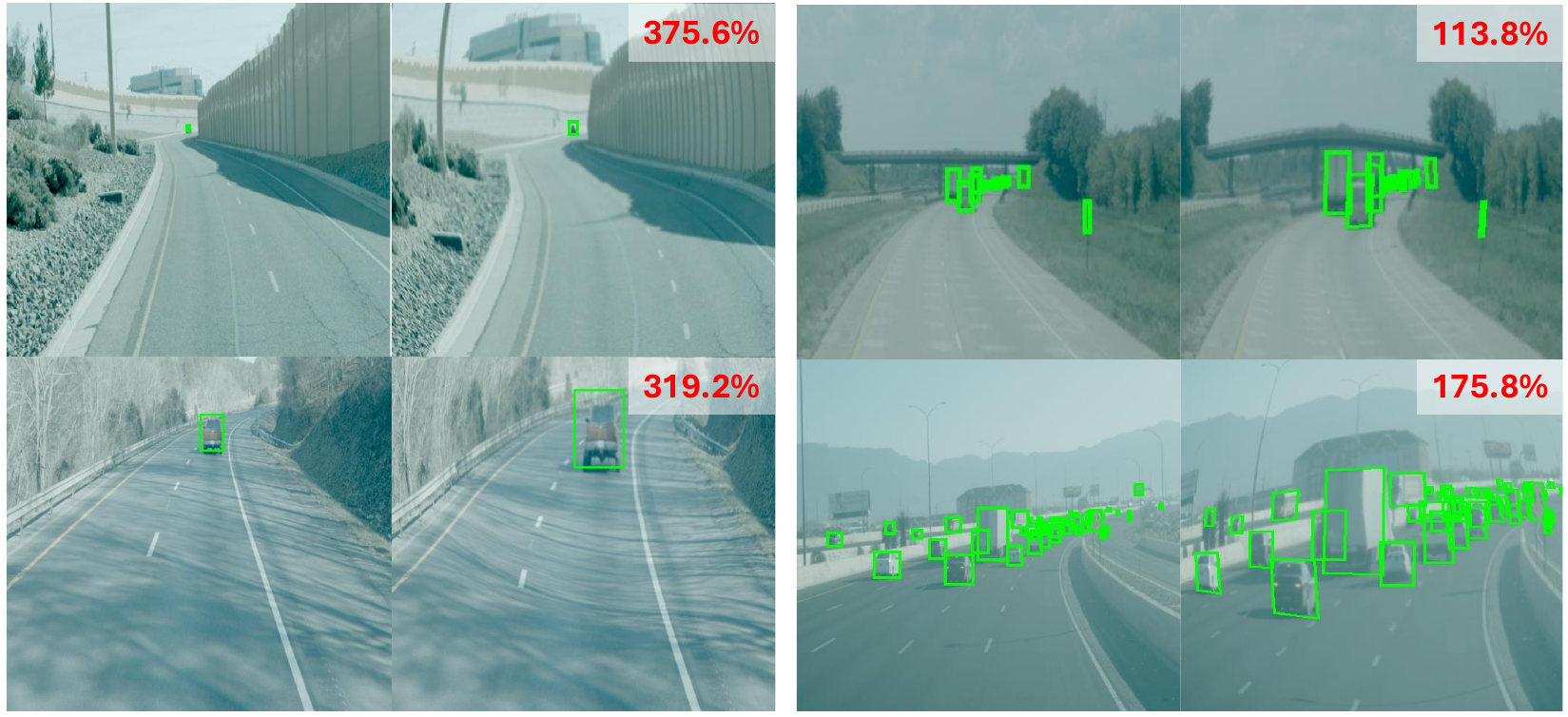}
    \caption{\textbf{Learned hyperbolic foveated transform on the TruckDrive dataset.} The original image (left) and the foveated image (right) are shown, together with the percentage increase in object bounding-box area. Both views are cropped to the same image region, highlighting the local magnification induced by the foveation transform. The proposed transform is effective for both isolated targets and dense, busy scenes.}
    \label{fig:foveated_examples}
\end{figure}

\subsection{Detection Network Architecture}\label{sec:network}

High-resolution images are needed to keep ultra long range objects resolvable.
As such, detection networks must scale favorably to image dimensions.
To this end, we propose a model architecture leveraging pre-trained foundation models with existing efficiencies baked in for optimal scaling performance.

Foundation model encoders have been shown to be strong object detection priors \cite{bolya2025perception, oquab2023dinov2, simeoni2025dinov3}.
A variety of foundation model encoders were tested with both DETR and Deformable DETR detection heads, and we find the optimal model combination to be the SAM3 image encoder and a Deformable DETR detection head.

Notably, the SAM3 image encoder uses windowed attention and sparse global attention to minimize the computational burden of processing high-resolution images \cite{carion2025sam}.
The Deformable DETR head shows fast convergence across all encoders and better evaluation metrics than DETR heads.
This is because DETR requires full self-attention at each decoder layer \cite{carion2020detr}, which distributes attention across the exponentially growing image patch features while Deformable DETR leverages sparse sampling \cite{zhu2020deformable}. \label{sec:method}
\section{Implementation}

\paragraph{Foveated Transform Parameters.}

The goal of the hyperbolic foveation is to magnify distant objects in the scene. To this end, we must first estimate where these objects are. We train a small FFN using the output of the image encoder to estimate the center, $o$, and radius, $R$, of the transform. A low-resolution image is used (i.e., $256\times256$ or $512\times512$) such that the parameter estimation incurs minimal computational overhead.

To set the other foveation parameter, we empirically determine which values maximize the transformed bounding boxes via grid search. For TruckDrive, the optimal parameters are set as $\alpha=2.0$ and $p=2.0$.

To compute the hyperbolic embeddings, we additionally train another FFN to project the set of foveation parameters into the same dimension as the object queries. This provides the detection network with information about the image-specific hyperbolic transformation such that the hyperbolic box parameterization, $b'$, can be estimated.
Examples of this learned transform are shown in Figure \ref{fig:foveated_examples}.

\paragraph{Training Losses.}
The hyperbolic box parameterization represents bounding boxes in the induced Riemannian space.
Unfortunately, defining distance and area in this space is non-trivial, making computing the losses between the target and predicted boxes challenging in Riemmanian space.

To overcome this challenge, the predicted boxes are projected into Euclidean space where vanilla L1 and gIoU losses are computed.
Notably, the iterative inverse \eqref{eq:inv_iter},  \eqref{eq:tangent_x}, and \eqref{eq:tangent_y} are all differentiable and are thus amenable to backpropagation.
The gIoU metric \cite{rezatofighi2019generalized} is better suited for detection of ultra-long range objects as it provides a gradient even when matched boxes do not overlap \cite{cheng2023survey}.
See Figure \ref{fig:iou_vs_giou} for a visualization of this phenomenon.

Following \cite{zhang2022dino}, we apply the de-noising training scheme, where ground truth boxes are noised and concatenated to the prediction queries to more effectively learn box alignment.
In this case the projected Riemannian boxes are used as the ground truth anchors to align them with the predictions in Riemmanian space. \label{sec:impl}
\section{Experimental Validation}
We evaluate the proposed ultra-long range detection framework on the TruckDrive dataset~\cite{ghilotti2026truckdrive}, and conduct controlled ablations to isolate the impact of backbone encoders, detection heads, and the proposed hyperbolic foveated transform.
Our evaluation focuses in particular on performance at long and ultra-long ranges, where existing detection pipelines degrade most severely.

\paragraph{Dataset and Metrics.}
As we are concerned with long and ultra-long range object detection, we use the TruckDrive dataset which is currently the only dataset with annotations at ultra-long range distances.
Since these ranges exceed the reliable operating regime of LiDAR and radar, object distances are estimated from bounding box height, camera intrinsics, and class-specific average object heights. 
An image resolution of $1024\times1024$ is used for all experiments.
We follow standard object detection protocols and report COCO-style mean average precision (mAP), together with distance-wise mAP computed over four distance bins
(\text{mAP}\textsubscript{0--50m}, \text{mAP}\textsubscript{50--150m}, \text{mAP}\textsubscript{150--250m}, and \text{mAP}\textsubscript{$>$250m}).
We additionally report PASCAL-style mAP at IoU thresholds of 0.5 and 0.75.

\begin{table}[t] 
\centering
\caption{\textbf{Ablation Experiments on the Components of Telescope}. A de-noising training scheme helps improve model performance and the hyperbolic foveation improves mAP at far, long, and ultra-long ranges while slightly reducing performance for nearby objects. All rows use SAM3 encoder.}
\label{table:telescope_ablation}
\resizebox{\linewidth}{!}{
\begin{tabular}{l|ccccc}
\toprule
\multirow{2}{*}{Method} & \multicolumn{5}{c}{COCO} \\
& mAP  & mAP\textsubscript{0-50} & mAP\textsubscript{50-150} & mAP\textsubscript{150-250} & mAP\textsubscript{250+} \\
\midrule \midrule
Deformable DETR & 0.32 & 0.52 & 0.34 & 0.24 & 0.17 \\
+ De-noising & 0.50 (\textbf{+0.18}) & 0.69 (\textbf{+0.17}) & 0.48 (\textbf{+0.14}) & 0.32 (\textbf{+0.08}) & 0.29 (\textbf{+0.12}) \\
+ Hyperbolic Foveation & 0.50 (\textbf{+0.00}) & 0.61 (\textbf{-0.08}) & 0.50 (\textbf{+0.02}) & 0.34 (\textbf{+0.02}) & 0.33 (\textbf{+0.03}) \\
\bottomrule
\end{tabular}
}
\end{table}
\begin{figure}
    \centering
    \includegraphics[width=\linewidth]{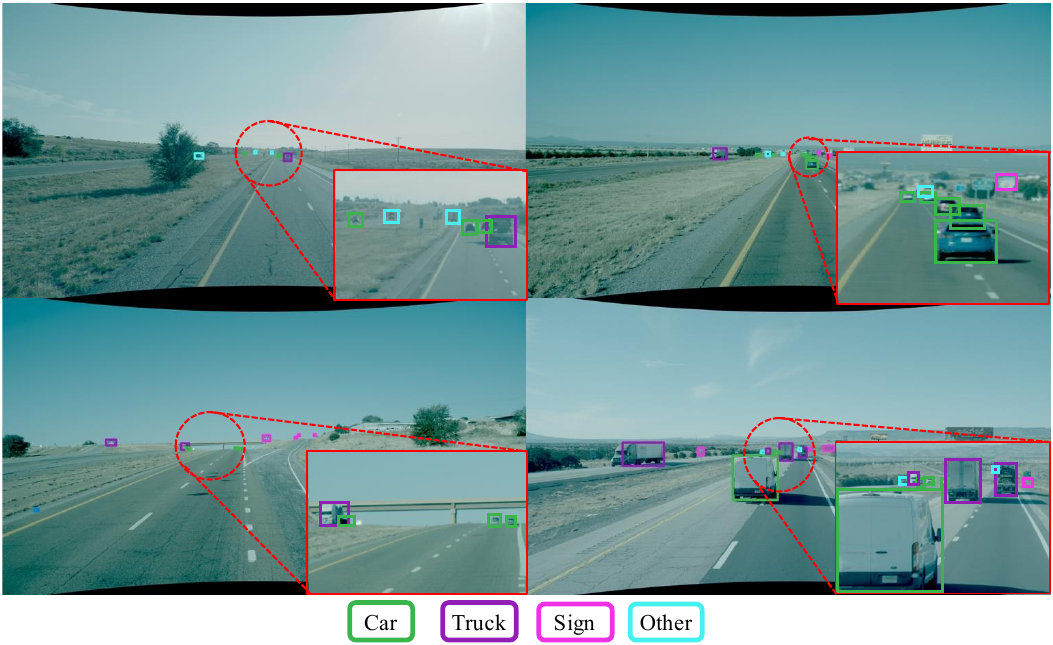}
    \caption{\textbf{Qualitative Visualization.} Detections from Telescope on the TruckDrive \cite{ghilotti2026truckdrive} dataset. Telescope consistently detects and localizes distant vehicles that occupy only a few pixels, while preserving accurate predictions for nearby objects. These examples highlight the effect of the proposed hyperbolic foveated transform in magnifying ultra-long range regions and improving sensitivity to objects in these ranges. Cut-outs provide a high-resolution, zoomed-in region to more clearly visualize objects.}
    \label{fig:zoomed_in}
\end{figure}
\subsection{Architecture Ablation Experiments}
\label{subsec:architecture}
\begin{table*}
\centering
\scriptsize
\caption{\textbf{Distance-wise Ablation Experiments of Encoder and Head}. All models perform well for objects close to the ego vehicle, but have degraded performance as distance increases. The proposed model using the SAM3 image encoder backbone, Deformable DETR detection head, and denoising-based training to provide optimal performance, especially at ultra-long range distances and across all classes.}
\label{table:ablation_distance_map}
\resizebox{0.8\linewidth}{!}{
\begin{tabular}{l|ccccc|cc}
\toprule
\multirow{2}{*}{Method} & \multicolumn{5}{c|}{COCO} & \multicolumn{2}{c}{PASCAL} \\
 & mAP & mAP\textsubscript{0-50} & mAP\textsubscript{50-150} & mAP\textsubscript{150-250} & mAP\textsubscript{250+} & mAP\textsubscript{50} & mAP\textsubscript{75} \\
\midrule \midrule
DINOv2 + Deformable DETR & 0.186 & 0.432 & 0.208 & 0.108 & 0.042 & 0.375 & 0.161 \\
DINOv3 + Deformable DETR & \cellcolor{tabthird} 0.212 & \cellcolor{tabthird} 0.467 & \cellcolor{tabthird} 0.250 & \cellcolor{tabthird} 0.137 & \cellcolor{tabthird} 0.059 & \cellcolor{tabthird} 0.419 & \cellcolor{tabthird} 0.190 \\
SAM3 + Deformable DETR & \cellcolor{tabsecond} 0.317 & \cellcolor{tabsecond} 0.523 & \cellcolor{tabsecond} 0.344 & \cellcolor{tabsecond} 0.244 & \cellcolor{tabsecond} 0.171 & \cellcolor{tabsecond} 0.521 & \cellcolor{tabsecond} 0.329 \\
SAM3 + Denoising~\cite{zhang2022dino} & \cellcolor{tabfirst} 0.501 & \cellcolor{tabfirst} 0.692 & \cellcolor{tabfirst} 0.483 & \cellcolor{tabfirst} 0.321 & \cellcolor{tabfirst} 0.292 & \cellcolor{tabfirst} 0.758 & \cellcolor{tabfirst} 0.545 \\
\bottomrule
\end{tabular}
}
\end{table*}
\begin{table}
\centering
\scriptsize
\caption{\textbf{Class-wise Ablation Experiments for Backbone and Head}. Results for the TruckDrive dataset for each class for existing state-of-the-art models as well as foundation model ablations. Leveraging foundation models as pre-trained image encoders provides a strong prior for object detection comparable to state-of-the-art specialized models. The proposed model using the SAM3 image encoder backbone, Deformable DETR detection head, and de-noising training scheme to provide optimal performance across all classes.}
\label{table:ablation_class_map}
\begin{tabularx}{\linewidth}{l|*{6}{>{\centering\arraybackslash}X}}
\toprule
\multirow{2}{*}{Method} & \multicolumn{6}{c}{COCO mAP} \\
 & Person & Bike & Sign & Car & Truck & Debris \\
\midrule \midrule
DINOv2 + Deformable DETR & \cellcolor{tabsecond} 0.154 & \cellcolor{tabthird} 0.302 & 0.175 & 0.302 & 0.301 & 0.092 \\
DINOv3 + Deformable DETR & \cellcolor{tabthird} 0.149 & \cellcolor{tabsecond} 0.304 & \cellcolor{tabthird} 0.222 & \cellcolor{tabthird} 0.345 & \cellcolor{tabthird} 0.334 & \cellcolor{tabthird} 0.120 \\
SAM3 + Deformable DETR & 0.106 & 0.236 & \cellcolor{tabsecond} 0.415 & \cellcolor{tabsecond} 0.459 & \cellcolor{tabsecond} 0.442 & \cellcolor{tabsecond} 0.245 \\
SAM3 + Denoising~\cite{zhang2022dino} & \cellcolor{tabfirst} 0.330 & \cellcolor{tabfirst} 0.436 & \cellcolor{tabfirst} 0.617 & \cellcolor{tabfirst} 0.631 & \cellcolor{tabfirst} 0.596 & \cellcolor{tabfirst} 0.451 \\
\bottomrule
\end{tabularx}
\end{table}
This study provides an analysis of which foundation model backbones and training schemes are most effective for ultra-long range detection.

Table~\ref{table:telescope_ablation} presents the ablation of the proposed Telescope pipeline.
Absolute improvements are highlighted in bold.
Notably, the de-noising training scheme significantly improves overall accuracy across all distance bins, confirming the utility of this approach for fine-tuning foundation models \cite{liu2024grounding}.
While the foveated transform slightly degrades performance at short distances, it consistently improves performance in the medium, long, and ultra-long regimes.
Importantly, the proposed transform significantly reduces the performance gap between near and far distance bins, producing a more balanced detector across spatial scales.
This behavior reflects the design goal of hyperbolic foveation, which magnifies distant objects and compresses nearby ones.

We next study the influence of the image encoder, detection head, and training scheme on model performance.
We evaluate five foundation model encoders, SAM3 \cite{carion2025sam}, DINOv2 \cite{oquab2023dinov2}, DINOv3 \cite{simeoni2025dinov3}, and Perception Encoder \cite{bolya2025perception}, in combination with both DETR \cite{carion2020detr} and Deformable DETR \cite{zhu2020deformable} heads.
To ensure a controlled comparison, all encoders are frozen and only the detection heads are trained from scratch for 12 epochs using identical hyperparameters. The results are summarized in Table~\ref{table:ablation_distance_map}. 
DETR-based models and models using the Perception Encoder consistently fail to converge and are therefore omitted from further ablation analysis. 
In contrast, Deformable DETR yields stable training and substantially better performance across all distance ranges.
Interestingly, the Perception Encoder backbone, despite being used internally by SAM3, also fails to converge for both DETR and Deformable DETR heads when used directly under the same training protocol.
\begin{figure*}
    \centering
    \includegraphics[width=\linewidth]{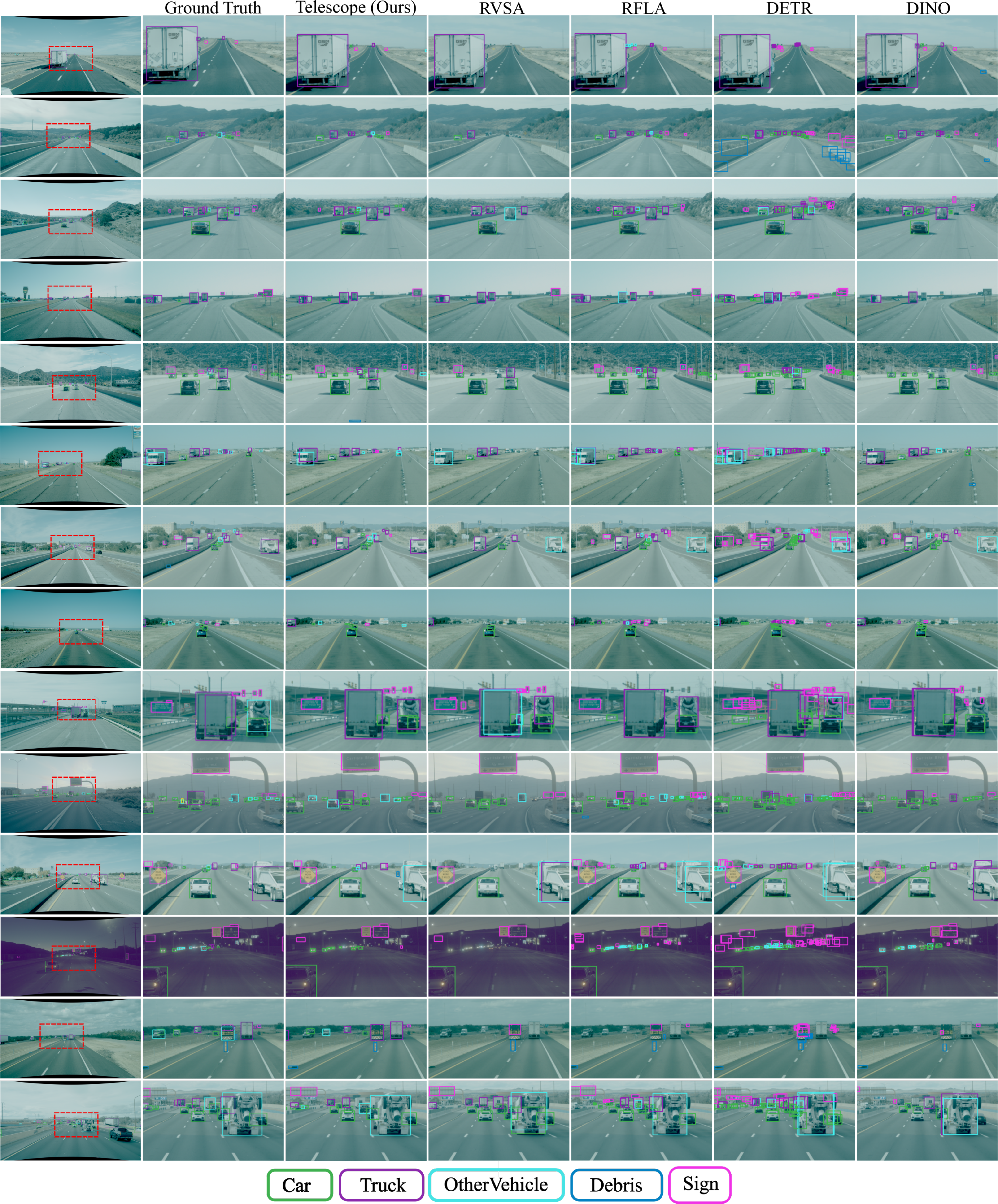}
    \caption{\textbf{Qualitative Comparison.} Qualitative comparison between the proposed method, Telescope, and state-of-the-art baselines. Both RVSA \cite{wang2022advancing} and RFLA \cite{xu2022rfla} are specialized for small object detection while DETR \cite{carion2020detr} and DINO \cite{zhang2022dino} are strong general object detectors, but perform worse in long an ultra-long range object detection. Ground truth annotations are shown on the left. Zoomed-in views corresponding to the red rectangles are provided to highlight detections at long and ultra-long range, where some objects reach up to $1$km. All baselines are fine-tuned on the TruckDrive \cite{ghilotti2026truckdrive} dataset.}
    \label{fig:full_page}
\end{figure*}
\begin{table*}
\centering
\scriptsize
\caption{\textbf{Distance-Wise Ultra-long Range Object Detection Evaluation on TruckDrive dataset}. 
The proposed model using the SAM3 image encoder backbone, Deformable DETR detection head, de-noising training, and TeleScope re-sampling layer provides optimal performance, especially at ultra long range distances, where it significantly improves over previous methods.}
\label{table:distance_map}
\resizebox{0.75\linewidth}{!}{
\begin{tabular}{l|ccccc|cc}
\toprule
\multirow{2}{*}{Method} & \multicolumn{5}{c|}{COCO} & \multicolumn{2}{c}{PASCAL} \\
 & mAP & mAP\textsubscript{0-50} & mAP\textsubscript{50-150} & mAP\textsubscript{150-250} & mAP\textsubscript{250+} & mAP\textsubscript{50} & mAP\textsubscript{75} \\
\midrule \midrule
DETR~\cite{carion2020detr} & 0.166 & 0.396 & 0.178 & 0.081 & 0.072 & 0.335 & 0.147 \\
Grounding DINO~\cite{liu2024grounding} & 0.286 & 0.376 & 0.262 & 0.147 & 0.156 & 0.417 & 0.296 \\
FOVEA~\cite{thavamani2021fovea} & 0.113 & 0.169 & 0.086 & 0.008 & 0.005 & 0.189 & 0.115 \\
YOLO11x~\cite{hidayatullah2025yolov8} & 0.266 & 0.421 & 0.218 & 0.134 & 0.117 & \cellcolor{tabthird} 0.510 & 0.195 \\
DINO~\cite{zhang2022dino} & 0.222 & 0.335 & 0.239 & 0.189 & 0.179 & 0.371 & 0.226 \\
QueryDet \cite{yang2022querydet} & 0.248 & 0.449 & 0.286 & 0.199 & 0.094 & 0.415 & 0.257 \\
UniverseNet \cite{shinya2021usb} & 0.305 & \cellcolor{tabsecond} 0.518 & \cellcolor{tabsecond} 0.334 & \cellcolor{tabthird} 0.236 & 0.145  & 0.474 & \cellcolor{tabthird} 0.318 \\
RVSA \cite{wang2022advancing} & \cellcolor{tabsecond} 0.325 & \cellcolor{tabthird} 0.502 & 0.298 & 0.233 & \cellcolor{tabthird} 0.183 & 0.488 & \cellcolor{tabsecond} 0.351 \\
RFLA \cite{xu2022rfla} & \cellcolor{tabthird} 0.306 & 0.501 & \cellcolor{tabthird} 0.320 & \cellcolor{tabsecond} 0.239 & \cellcolor{tabsecond} 0.185 & \cellcolor{tabsecond} 0.512 & 0.317 \\
Telescope (Ours) & \cellcolor{tabfirst} 0.497 & \cellcolor{tabfirst} 0.608 & \cellcolor{tabfirst} 0.507 & \cellcolor{tabfirst} 0.335 & \cellcolor{tabfirst} 0.326 & \cellcolor{tabfirst} 0.801 & \cellcolor{tabfirst} 0.494 \\
\bottomrule
\end{tabular}
}
\end{table*}
\begin{table}
\centering
\scriptsize
\caption{\textbf{Class-Wise Ultra-Long Range Detection Evaluation on TruckDrive dataset}. Leveraging foundation models as pre-trained image encoders provides a strong prior for object detection comparable to state-of-the-art specialized models. The proposed model using the SAM3 image encoder backbone, Deformable DETR detection head, de-noising training, and TeleScope re-sampling layer provides optimal performance across all classes.}
\label{table:class_map}
\begin{tabularx}{\linewidth}{l|*{6}{>{\centering\arraybackslash}X}}
\toprule
\multirow{2}{*}{Method} &\multicolumn{6}{c}{COCO mAP} \\
 & Person & Bike & Sign & Car & Truck & Debris \\\midrule \midrule
DETR \cite{carion2020detr} & \cellcolor{tabsecond} 0.222 & 0.327 & 0.179 & 0.299 & 0.247 & 0.083 \\
Grounding DINO~\cite{liu2024grounding} & 0.141 & 0.174 & 0.472 & \cellcolor{tabsecond} 0.591 & 0.317 & 0.024 \\
FOVEA~\cite{thavamani2021fovea} & 0.028 & 0.043 & 0.060 & 0.377 & 0.056 & 0.107 \\
YOLO11x \cite{hidayatullah2025yolov8} & \cellcolor{tabthird} 0.172 & \cellcolor{tabthird} 0.400 & 0.370 & 0.330 & 0.376 & 0.049 \\
DINO \cite{zhang2022dino} & 0.059 & \cellcolor{tabfirst} 0.819 & 0.264 & 0.431 & 0.334 & \cellcolor{tabthird} 0.226 \\
QueryDet \cite{yang2022querydet} & 0.034 & 0.222 & 0.329 & 0.469 & 0.349 & 0.185 \\
UniverseNet \cite{shinya2021usb} & 0.165 & 0.240 & \cellcolor{tabthird} 0.429 & 0.538 & 0.427 & \cellcolor{tabsecond} 0.230 \\
RVSA \cite{wang2022advancing} & 0.069 & 0.297 & \cellcolor{tabthird} 0.429 & 0.551 & \cellcolor{tabsecond} 0.467 & 0.203 \\
RFLA \cite{xu2022rfla} & 0.085 & 0.209 & \cellcolor{tabsecond} 0.436 & \cellcolor{tabthird} 0.560 & \cellcolor{tabthird} 0.434 & 0.225 \\
Telescope (Ours) & \cellcolor{tabfirst} 0.454 & \cellcolor{tabsecond} 0.620 & \cellcolor{tabfirst} 0.568 & \cellcolor{tabfirst} 0.651 & \cellcolor{tabfirst} 0.595 & \cellcolor{tabfirst} 0.397 \\
\bottomrule
\end{tabularx}
\vspace{10pt}
\end{table}

Among the tested encoders, SAM3 provides the strongest overall performance.
This reflects the strong spatial inductive biases inherited from large-scale segmentation pre-training.
While DINOv2 and DINOv3 achieve competitive results for several categories, their overall accuracy remains lower under the same compute budget. Table~\ref{table:ablation_class_map} further shows the class-dependent behavior across foundation encoders.
DINO-based encoders tend to perform slightly better on pedestrians and bicycles, whereas SAM-based encoders favor vehicles and traffic signs.
As the application of this ultra-long range object detection network is autonomous highway driving, this motivates the use of the SAM3 backbone as it aligns with the application domain.
Based on these results, we adopt the SAM3 encoder with a Deformable DETR head in all subsequent experiments.

We evaluate the denoising training scheme~\cite{zhang2022dino} and associated losses using the SAM3 image encoder.
As shown in Tables \ref{table:telescope_ablation} \ref{table:ablation_distance_map}, and \ref{table:ablation_distance_map}, de-noising improves detection accuracy across all distance bins and classes and is therefore adopted within the Telescope model. 

Figure \ref{fig:zoomed_in} presents qualitative examples showing that the full 2-stage Telescope model accurately detects objects at long and ultra-long ranges.

\subsection{Ultra Long Range Object Detection} \label{subsec:object_det_models}

We next compare Telescope against state-of-the-art 2D object detectors in Tables~\ref{table:distance_map} and~\ref{table:class_map}.
All baselines are initialized from the best publicly available checkpoints and fine-tuned on TruckDrive for 12 epochs following \cite{ghilotti2026truckdrive}.
Telescope is trained following the same protocol used in the ablation study described in Section~\ref{subsec:architecture}.
Notably, QueryDet \cite{yang2022querydet}, UniverseNet \cite{shinya2021usb}, RVSA \cite{wang2022advancing}, FOVEA \cite{thavamani2021fovea}, and RFLA \cite{xu2022rfla} are all designed for small-object detection, while Grounding DINO \cite{liu2024grounding} is a state-of-the-art visual-language model.

Baselines in Tables \ref{table:distance_map} and \ref{table:class_map} rely on backbones pre-trained on relatively modest datasets (e.g., ImageNet \cite{deng2009imagenet}) and require small-object–specific losses, data augmentations, and specialized training strategies. 
In contrast, Tables \ref{table:ablation_distance_map} and \ref{table:ablation_class_map} show that simply initializing from a stronger foundation encoder (SAM3), trained at much larger scale, and applying standard training on modest dataset already matches or exceeds these specialized methods. 
Incorporating de-noising and foveation further yields substantial gains, clearly outperforming the strongest existing approaches by a wide margin as seen in Table \ref{table:distance_map}.

These results demonstrate that explicitly re-balancing object scales through hyperbolic foveation, together with a pre-trained image encoder and de-noising training, improves sensitivity to distant objects with an mAP increase of $76\%$ for ultra-long range (increasing it from 0.185
to 0.326 for distances greater than $250\mathrm{m}$) without sacrificing overall detection quality.

A qualitative comparison between the proposed method, the strongest small-object detection baselines, and widely used general object detectors is shown in Figure~\ref{fig:full_page}.
DETR, DINO, and RFLA tend to over-predict the number of objects in a scene, whereas RVSA is more conservative and produces fewer detections.
Telescope offers a middle ground, reducing false positives relative to DETR and DINO while maintaining higher recall than RVSA.

Additional medium and long-range ($<250$m) experiments on the Argoverse \cite{chang2019argoverse, wilson2023argoverse} dataset are presented in the Appendix. \label{sec:exp}
\section{Conclusion}

We present Telescope, a two-stage algorithm for ultra-long range object detection that explicitly addresses the extreme scale imbalance inherent in autonomous highway driving scenarios.
In the first stage, we introduce a learnable hyperbolic foveated transform that magnifies distant regions while compressing nearby ones, normalizing object scales and reducing the dominance of large, nearby objects.
In the second stage, we combine this transformation with a high-resolution detection architecture built on a foundation model image encoder and Deformable DETR detection head, enabling efficient processing and training without the quadratic cost of standard self-attention.

Experiments on the long-range TruckDrive dataset demonstrate that the proposed foveation consistently improves detection accuracy for distant objects and reduces the performance gap between near and far ranges.
In particular, Telescope achieves a 53\% relative improvement in overall performance (increasing overall mAP from 0.325 to 0.497) but most notably achieves up to a 76\% relative improvement in mAP over the strongest existing baselines at ultra-long distances, increasing absolute mAP from 0.185 to 0.326 at distances greater than $250\mathrm{m}$.

We note that the proposed hyperbolic foveated transform is architecture-agnostic and invertible, and can be readily integrated into existing high-resolution perception pipelines, including future multi-modal and vision–language detection systems. We believe this work establishes a principled and extensible foundation for addressing the representational challenges of simultaneous perception at close surroundings with up beyond hundreds of meters to kilometer-scale distances. \label{sec:conc}

\section{Limitations and Scope}
Telescope is a research contribution to one component of a broader autonomous perception system. Deployment in safety-critical applications would require integration with complementary sensing modalities, system-level validation, and compliance with applicable regulatory frameworks. The results reported here reflect performance on the TruckDrive dataset and should not be interpreted as a guarantee of real-world system performance.

\section*{Acknowledgements}
Felix Heide was supported by an NSF CAREER Award (2047359), a Packard Foundation Fellowship, a Sloan Research Fellowship, a Sony Young Faculty Award, a Project X Innovation Award and a Amazon Science Research Award. Felix Heide is a co-founder of Algolux (now Torc Robotics), Head of AI at Torc Robotics, and a cofounder of Cephia AI.

\FloatBarrier
{
    \small
    \bibliographystyle{splncs04}
    \bibliography{main}

@inproceedings{geiger2012kitti,
  title={Are we ready for autonomous driving? {T}he {KITTI} vision benchmark suite},
  author={Geiger, Andreas and Lenz, Philip and Urtasun, Raquel},
  booktitle={2012 IEEE conference on computer vision and pattern recognition},
  pages={3354--3361},
  year={2012},
  organization={IEEE}
}

@inproceedings{caesar2020nuscenes,
  title={nu{S}cenes: {A} multimodal dataset for autonomous driving},
  author={Caesar, Holger and Bankiti, Varun and Lang, Alex H and Vora, Sourabh and Liong, Venice Erin and Xu, Qiang and Krishnan, Anush and Pan, Yu and Baldan, Giancarlo and Beijbom, Oscar},
  booktitle={Proceedings of the IEEE/CVF conference on computer vision and pattern recognition},
  pages={11621--11631},
  year={2020}
}

@inproceedings{sun2020waymo,
  title={Scalability in perception for autonomous driving: {W}aymo open dataset},
  author={Sun, Pei and Kretzschmar, Henrik and Dotiwalla, Xerxes and Chouard, Aurelien and Patnaik, Vijaysai and Tsui, Paul and Guo, James and Zhou, Yin and Chai, Yuning and Caine, Benjamin and others},
  booktitle={Proceedings of the IEEE/CVF conference on computer vision and pattern recognition},
  pages={2446--2454},
  year={2020}
}

@inproceedings{chang2019argoverse,
  title={Argoverse: 3{D} tracking and forecasting with rich maps},
  author={Chang, Ming-Fang and Lambert, John and Sangkloy, Patsorn and Singh, Jagjeet and Bak, Slawomir and Hartnett, Andrew and Wang, De and Carr, Peter and Lucey, Simon and Ramanan, Deva and others},
  booktitle={Proceedings of the IEEE/CVF conference on computer vision and pattern recognition},
  pages={8748--8757},
  year={2019}
}

@article{ren2015faster,
  title={Faster {R-CNN}: {T}owards real-time object detection with region proposal networks},
  author={Ren, Shaoqing and He, Kaiming and Girshick, Ross and Sun, Jian},
  journal={Advances in neural information processing systems},
  volume={28},
  year={2015}
}

@inproceedings{redmon2016yolo,
  title={You only look once: {U}nified, real-time object detection},
  author={Redmon, Joseph and Divvala, Santosh and Girshick, Ross and Farhadi, Ali},
  booktitle={Proceedings of the IEEE conference on computer vision and pattern recognition},
  pages={779--788},
  year={2016}
}

@inproceedings{carion2020detr,
  title={End-to-end object detection with transformers},
  author={Carion, Nicolas and Massa, Francisco and Synnaeve, Gabriel and Usunier, Nicolas and Kirillov, Alexander and Zagoruyko, Sergey},
  booktitle={European conference on computer vision},
  pages={213--229},
  year={2020},
  organization={Springer}
}

@article{zhu2020deformable,
  title={Deformable {DETR}: {D}eformable transformers for end-to-end object detection},
  author={Zhu, Xizhou and Su, Weijie and Lu, Lewei and Li, Bin and Wang, Xiaogang and Dai, Jifeng},
  journal={arXiv preprint arXiv:2010.04159},
  year={2020}
}

@article{zhou2019objects,
  title={Objects as points},
  author={Zhou, Xingyi and Wang, Dequan and Kr{\"a}henb{\"u}hl, Philipp},
  journal={arXiv preprint arXiv:1904.07850},
  year={2019}
}

@inproceedings{tian2019fcos,
  title={{FCOS}: {F}ully convolutional one-stage object detection},
  author={Tian, Zhi and Shen, Chunhua and Chen, Hao and He, Tong},
  booktitle={Proceedings of the IEEE/CVF international conference on computer vision},
  pages={9627--9636},
  year={2019}
}

@article{cheng2023survey,
  title={Towards large-scale small object detection: {S}urvey and benchmarks},
  author={Cheng, Gong and Yuan, Xiang and Yao, Xiwen and Yan, Kebing and Zeng, Qinghua and Xie, Xingxing and Han, Junwei},
  journal={IEEE transactions on pattern analysis and machine intelligence},
  volume={45},
  pages={13467--13488},
  year={2023},
  publisher={IEEE}
}

@inproceedings{yu2020scale,
  title={Scale match for tiny person detection},
  author={Yu, Xuehui and Gong, Yuqi and Jiang, Nan and Ye, Qixiang and Han, Zhenjun},
  booktitle={Proceedings of the IEEE/CVF winter conference on applications of computer vision},
  pages={1257--1265},
  year={2020}
}

@article{xu2022detecting,
  title={Detecting tiny objects in aerial images: {A} normalized {W}asserstein distance and a new benchmark},
  author={Xu, Chang and Wang, Jinwang and Yang, Wen and Yu, Huai and Yu, Lei and Xia, Gui-Song},
  journal={ISPRS Journal of Photogrammetry and Remote Sensing},
  volume={190},
  pages={79--93},
  year={2022},
  publisher={Elsevier}
}

@inproceedings{xia2018dota,
  title={{DOTA}: {A} large-scale dataset for object detection in aerial images},
  author={Xia, Gui-Song and Bai, Xiang and Ding, Jian and Zhu, Zhen and Belongie, Serge and Luo, Jiebo and Datcu, Mihai and Pelillo, Marcello and Zhang, Liangpei},
  booktitle={Proceedings of the IEEE conference on computer vision and pattern recognition},
  pages={3974--3983},
  year={2018}
}

@inproceedings{lin2017feature,
  title={Feature pyramid networks for object detection},
  author={Lin, Tsung-Yi and Doll{\'a}r, Piotr and Girshick, Ross and He, Kaiming and Hariharan, Bharath and Belongie, Serge},
  booktitle={Proceedings of the IEEE conference on computer vision and pattern recognition},
  pages={2117--2125},
  year={2017}
}

@inproceedings{bai2018finding,
  title={Finding tiny faces in the wild with generative adversarial network},
  author={Bai, Yancheng and Zhang, Yongqiang and Ding, Mingli and Ghanem, Bernard},
  booktitle={Proceedings of the IEEE conference on computer vision and pattern recognition},
  pages={21--30},
  year={2018}
}

@article{chen2020mmdetection,
  title   = {{MMDetection}: {O}pen {MML}ab Detection Toolbox and Benchmark},
  author  = {Chen, Kai and Wang, Jiaqi and Pang, Jiangmiao and Cao, Yuhang and
             Xiong, Yu and Li, Xiaoxiao and Sun, Shuyang and Feng, Wansen and
             Liu, Ziwei and Xu, Jiarui and Zhang, Zheng and Cheng, Dazhi and
             Zhu, Chenchen and Cheng, Tianheng and Zhao, Qijie and Li, Buyu and
             Lu, Xin and Zhu, Rui and Wu, Yue and Dai, Jifeng and Wang, Jingdong
             and Shi, Jianping and Ouyang, Wanli and Loy, Chen Change and Lin, Dahua},
  journal= {arXiv preprint arXiv:1906.07155},
  year={2019}
}

@inproceedings{thavamani2021fovea,
  title={{FOVEA}: {Fo}veated image magnification for autonomous navigation},
  author={Thavamani, Chittesh and Li, Mengtian and Cebron, Nicolas and Ramanan, Deva},
  booktitle={Proceedings of the IEEE/CVF international conference on computer vision},
  pages={15539--15548},
  year={2021}
}

@article{jaderberg2015spatial,
  title={Spatial transformer networks},
  author={Jaderberg, Max and Simonyan, Karen and Zisserman, Andrew and others},
  journal={Advances in neural information processing systems},
  volume={28},
  year={2015}
}

@article{jabbireddy2022foveated,
  title={Foveated rendering: {M}otivation, taxonomy, and research directions},
  author={Jabbireddy, Susmija and Sun, Xuetong and Meng, Xiaoxu and Varshney, Amitabh},
  journal={arXiv preprint arXiv:2205.04529},
  year={2022}
}

@article{yurtsever2020survey,
  title={A survey of autonomous driving: {C}ommon practices and emerging technologies},
  author={Yurtsever, Ekim and Lambert, Jacob and Carballo, Alexander and Takeda, Kazuya},
  journal={IEEE access},
  volume={8},
  pages={58443--58469},
  year={2020},
  publisher={IEEE}
}

@article{wong2020mapping,
  title={Mapping for autonomous driving: {O}pportunities and challenges},
  author={Wong, Kelvin and Gu, Yanlei and Kamijo, Shunsuke},
  journal={IEEE Intelligent Transportation Systems Magazine},
  volume={13},
  number={1},
  pages={91--106},
  year={2020},
  publisher={IEEE}
}

@inproceedings{huang2018apolloscape,
  title={The {A}polloscape dataset for autonomous driving},
  author={Huang, Xinyu and Cheng, Xinjing and Geng, Qichuan and Cao, Binbin and Zhou, Dingfu and Wang, Peng and Lin, Yuanqing and Yang, Ruigang},
  booktitle={Proceedings of the IEEE conference on computer vision and pattern recognition workshops},
  pages={954--960},
  year={2018}
}

@article{wilson2023argoverse,
  title={Argoverse 2: {N}ext generation datasets for self-driving perception and forecasting},
  author={Wilson, Benjamin and Qi, William and Agarwal, Tanmay and Lambert, John and Singh, Jagjeet and Khandelwal, Siddhesh and Pan, Bowen and Kumar, Ratnesh and Hartnett, Andrew and Pontes, Jhony Kaesemodel and others},
  journal={arXiv preprint arXiv:2301.00493},
  year={2023}
}

@book{lee2018introduction,
  title={Introduction to {R}iemannian manifolds},
  author={Lee, John M},
  volume={2},
  year={2018},
  publisher={Springer}
}

@article{ravi2024sam,
  title={{SAM} 2: {S}egment anything in images and videos},
  author={Ravi, Nikhila and Gabeur, Valentin and Hu, Yuan-Ting and Hu, Ronghang and Ryali, Chaitanya and Ma, Tengyu and Khedr, Haitham and R{\"a}dle, Roman and Rolland, Chloe and Gustafson, Laura et al.},
  journal={arXiv preprint arXiv:2408.00714},
  year={2024}
}

@article{carion2025sam,
  title={{SAM} 3: {S}egment anything with concepts},
  author={Carion, Nicolas and Gustafson, Laura and Hu, Yuan-Ting and Debnath, Shoubhik and Hu, Ronghang and Suris, Didac and Ryali, Chaitanya and Alwala, Kalyan Vasudev and Khedr, Haitham and Huang, Andrew et al.},
  journal={arXiv preprint arXiv:2511.16719},
  year={2025}
}

@article{oquab2023dinov2,
  title={{DINO}v2: {L}earning robust visual features without supervision},
  author={Oquab, Maxime and Darcet, Timoth{\'e}e and Moutakanni, Th{\'e}o and Vo, Huy and Szafraniec, Marc and Khalidov, Vasil and Fernandez, Pierre and Haziza, Daniel and Massa, Francisco and El-Nouby, Alaaeldin and others},
  journal={arXiv preprint arXiv:2304.07193},
  year={2023}
}

@article{simeoni2025dinov3,
  title={{DINO}v3},
  author={Sim{\'e}oni, Oriane and Vo, Huy V and Seitzer, Maximilian and Baldassarre, Federico and Oquab, Maxime and Jose, Cijo and Khalidov, Vasil and Szafraniec, Marc and Yi, Seungeun and Ramamonjisoa, Micha{\"e}l and others},
  journal={arXiv preprint arXiv:2508.10104},
  year={2025}
}

@article{bolya2025perception,
  title={Perception {E}ncoder: {T}he best visual embeddings are not at the output of the network},
  author={Bolya, Daniel and Huang, Po-Yao and Sun, Peize and Cho, Jang Hyun and Madotto, Andrea and Wei, Chen and Ma, Tengyu and Zhi, Jiale and Rajasegaran, Jathushan and Rasheed, Hanoona and others},
  journal={arXiv preprint arXiv:2504.13181},
  year={2025}
}

@article{zou2023object,
  title={Object detection in 20 years: {A} survey},
  author={Zou, Zhengxia and Chen, Keyan and Shi, Zhenwei and Guo, Yuhong and Ye, Jieping},
  journal={Proceedings of the IEEE},
  volume={111},
  number={3},
  pages={257--276},
  year={2023},
  publisher={IEEE}
}

@article{zhao2019object,
  title={Object detection with deep learning: {A} review},
  author={Zhao, Zhong-Qiu and Zheng, Peng and Xu, Shou-tao and Wu, Xindong},
  journal={IEEE transactions on neural networks and learning systems},
  volume={30},
  number={11},
  pages={3212--3232},
  year={2019},
  publisher={IEEE}
}

@article{mao20233d,
  title={3{D} object detection for autonomous driving: {A} comprehensive survey},
  author={Mao, Jiageng and Shi, Shaoshuai and Wang, Xiaogang and Li, Hongsheng},
  journal={International Journal of Computer Vision},
  volume={131},
  number={8},
  pages={1909--1963},
  year={2023},
  publisher={Springer}
}

@inproceedings{gong2021effective,
  title={Effective fusion factor in {FPN} for tiny object detection},
  author={Gong, Yuqi and Yu, Xuehui and Ding, Yao and Peng, Xiaoke and Zhao, Jian and Han, Zhenjun},
  booktitle={Proceedings of the IEEE/CVF winter conference on applications of computer vision},
  pages={1160--1168},
  year={2021}
}

@article{guo2023save,
  title={Save the tiny, save the all: {H}ierarchical activation network for tiny object detection},
  author={Guo, Guangqian and Chen, Pengfei and Yu, Xuehui and Han, Zhenjun and Ye, Qixiang and Gao, Shan},
  journal={IEEE transactions on circuits and systems for video technology},
  volume={34},
  number={1},
  pages={221--234},
  year={2023},
  publisher={IEEE}
}

@article{wang2021normalized,
  title={A normalized {G}aussian {W}asserstein distance for tiny object detection},
  author={Wang, Jinwang and Xu, Chang and Yang, Wen and Yu, Lei},
  journal={arXiv preprint arXiv:2110.13389},
  year={2021}
}

@inproceedings{de2023unbalanced,
  title={Unbalanced optimal transport: {A} unified framework for object detection},
  author={De Plaen, Henri and De Plaen, Pierre-Fran{\c{c}}ois and Suykens, Johan AK and Proesmans, Marc and Tuytelaars, Tinne and Van Gool, Luc},
  booktitle={Proceedings of the IEEE/CVF Conference on Computer Vision and Pattern Recognition},
  pages={3198--3207},
  year={2023}
}

@article{zhang2022dino,
  title={{DINO}: {DETR} with improved denoising anchor boxes for end-to-end object detection},
  author={Zhang, Hao and Li, Feng and Liu, Shilong and Zhang, Lei and Su, Hang and Zhu, Jun and Ni, Lionel M and Shum, Heung-Yeung},
  journal={arXiv preprint arXiv:2203.03605},
  year={2022}
}

@article{hidayatullah2025yolov8,
  title={{YOLO}v8 to {YOLO}11: {A} comprehensive architecture in-depth comparative review},
  author={Hidayatullah, Priyanto and Syakrani, Nurjannah and Sholahuddin, Muhammad Rizqi and Gelar, Trisna and Tubagus, Refdinal},
  journal={arXiv preprint arXiv:2501.13400},
  year={2025}
}

@inproceedings{cordts2016cityscapes,
  title={The cityscapes dataset for semantic urban scene understanding},
  author={Cordts, Marius and Omran, Mohamed and Ramos, Sebastian and Rehfeld, Timo and Enzweiler, Markus and Benenson, Rodrigo and Franke, Uwe and Roth, Stefan and Schiele, Bernt},
  booktitle={Proceedings of the IEEE conference on computer vision and pattern recognition},
  pages={3213--3223},
  year={2016}
}

@article{wang2022advancing,
  title={Advancing plain vision transformer toward remote sensing foundation model},
  author={Wang, Di and Zhang, Qiming and Xu, Yufei and Zhang, Jing and Du, Bo and Tao, Dacheng and Zhang, Liangpei},
  journal={IEEE transactions on geoscience and remote sensing},
  volume={61},
  pages={1--15},
  year={2022},
  publisher={IEEE}
}

@inproceedings{xu2022rfla,
  title={RFLA: Gaussian receptive field based label assignment for tiny object detection},
  author={Xu, Chang and Wang, Jinwang and Yang, Wen and Yu, Huai and Yu, Lei and Xia, Gui-Song},
  booktitle={European conference on computer vision},
  pages={526--543},
  year={2022},
  organization={Springer}
}

@article{shinya2021usb,
  title={{USB}: {U}niversal-scale object detection benchmark},
  author={Shinya, Yosuke},
  journal={arXiv preprint arXiv:2103.14027},
  year={2021}
}

@inproceedings{yang2022querydet,
  title={Query{D}et: {C}ascaded sparse query for accelerating high-resolution small object detection},
  author={Yang, Chenhongyi and Huang, Zehao and Wang, Naiyan},
  booktitle={Proceedings of the IEEE/CVF Conference on computer vision and pattern recognition},
  pages={13668--13677},
  year={2022}
}

@article{wei2024review,
  title={A review of small object detection based on deep learning},
  author={Wei, Wei and Cheng, Yu and He, Jiafeng and Zhu, Xiyue},
  journal={Neural Computing and Applications},
  volume={36},
  number={12},
  pages={6283--6303},
  year={2024},
  publisher={Springer}
}

@article{mirzaei2023small,
  title={Small object detection and tracking: {A} comprehensive review},
  author={Mirzaei, Behzad and Nezamabadi-Pour, Hossein and Raoof, Amir and Derakhshani, Reza},
  journal={Sensors},
  volume={23},
  number={15},
  pages={6887},
  year={2023},
  publisher={MDPI}
}

@inproceedings{chen2016r,
  title={{R-CNN} for small object detection},
  author={Chen, Chenyi and Liu, Ming-Yu and Tuzel, Oncel and Xiao, Jianxiong},
  booktitle={Asian conference on computer vision},
  pages={214--230},
  year={2016},
  organization={Springer}
}

@inproceedings{xia2022vision,
  title={Vision transformer with deformable attention},
  author={Xia, Zhuofan and Pan, Xuran and Song, Shiji and Li, Li Erran and Huang, Gao},
  booktitle={Proceedings of the IEEE/CVF conference on computer vision and pattern recognition},
  pages={4794--4803},
  year={2022}
}

@inproceedings{dai2021dynamic,
  title={Dynamic {DETR}: {E}nd-to-end object detection with dynamic attention},
  author={Dai, Xiyang and Chen, Yinpeng and Yang, Jianwei and Zhang, Pengchuan and Yuan, Lu and Zhang, Lei},
  booktitle={Proceedings of the IEEE/CVF international conference on computer vision},
  pages={2988--2997},
  year={2021}
}

@inproceedings{li2022dn,
  title={{DN-DETR}: {A}ccelerate {DETR} training by introducing query denoising},
  author={Li, Feng and Zhang, Hao and Liu, Shilong and Guo, Jian and Ni, Lionel M and Zhang, Lei},
  booktitle={Proceedings of the IEEE/CVF conference on computer vision and pattern recognition},
  pages={13619--13627},
  year={2022}
}

@inproceedings{dai2021up,
  title={{UP-DETR}: {U}nsupervised pre-training for object detection with transformers},
  author={Dai, Zhigang and Cai, Bolun and Lin, Yugeng and Chen, Junying},
  booktitle={Proceedings of the IEEE/CVF conference on computer vision and pattern recognition},
  pages={1601--1610},
  year={2021}
}

@article{nguyen2020evaluation,
  title={An evaluation of deep learning methods for small object detection},
  author={Nguyen, Nhat-Duy and Do, Tien and Ngo, Thanh Duc and Le, Duy-Dinh},
  journal={Journal of electrical and computer engineering},
  volume={2020},
  number={1},
  pages={3189691},
  year={2020},
  publisher={Wiley Online Library}
}

@article{lee2021self,
  title={Self-supervised feature enhancement networks for small object detection in noisy images},
  author={Lee, Geonsoo and Hong, Sungeun and Cho, Donghyeon},
  journal={IEEE signal processing letters},
  volume={28},
  pages={1026--1030},
  year={2021},
  publisher={IEEE}
}

@inproceedings{noh2019better,
  title={Better to follow, follow to be better: {T}owards precise supervision of feature super-resolution for small object detection},
  author={Noh, Junhyug and Bae, Wonho and Lee, Wonhee and Seo, Jinhwan and Kim, Gunhee},
  booktitle={Proceedings of the IEEE/CVF international conference on computer vision},
  pages={9725--9734},
  year={2019}
}

@article{liu2024small,
  title={Small-object detection in remote sensing images with super-resolution perception},
  author={Liu, Jiahang and Zhang, Jinlong and Ni, Yue and Chi, Weijian and Qi, Zitong},
  journal={IEEE Journal of Selected Topics in Applied Earth Observations and Remote Sensing},
  volume={17},
  pages={15721--15734},
  year={2024},
  publisher={IEEE}
}

@inproceedings{liu2024grounding,
  title={Grounding {DINO}: {M}arrying {DINO} with grounded pre-training for open-set object detection},
  author={Liu, Shilong and Zeng, Zhaoyang and Ren, Tianhe and Li, Feng and Zhang, Hao and Yang, Jie and Jiang, Qing and Li, Chunyuan and Yang, Jianwei and Su, Hang and others},
  booktitle={European conference on computer vision},
  pages={38--55},
  year={2024},
  organization={Springer}
}

@inproceedings{ghilotti2026truckdrive,
  title={Truck{D}rive: {L}ong-Range Autonomous Highway Driving Dataset},
  author={Ghilotti, Filippo and Palladin, Edoardo and Brucker, Samuel and Sigal, Adam and Bijelic, Mario and Heide, Felix},
  booktitle={Proceedings of the IEEE/CVF Conference on Computer Vision and Pattern Recognition},
  pages={},
  year={2026}
}

@inproceedings{rezatofighi2019generalized,
  title={Generalized intersection over union: {A} metric and a loss for bounding box regression},
  author={Rezatofighi, Hamid and Tsoi, Nathan and Gwak, JunYoung and Sadeghian, Amir and Reid, Ian and Savarese, Silvio},
  booktitle={Proceedings of the IEEE/CVF conference on computer vision and pattern recognition},
  pages={658--666},
  year={2019}
}

@article{stanoyevitch1994geometry,
  title={The geometry of {P}oincar{\'e} disks},
  author={Stanoyevitch, Alexander and Stegenga, David A},
  journal={Complex Variables and Elliptic Equations},
  volume={24},
  number={3-4},
  pages={249--265},
  year={1994},
  publisher={Taylor \& Francis}
}

@inproceedings{deng2009imagenet,
  title={Imagenet: {A} large-scale hierarchical image database},
  author={Deng, Jia and Dong, Wei and Socher, Richard and Li, Li-Jia and Li, Kai and Fei-Fei, Li},
  booktitle={2009 IEEE conference on computer vision and pattern recognition},
  pages={248--255},
  year={2009},
  organization={Ieee}
}

@inproceedings{zhao2024simple,
  title={Simple-{FPN}: {A}n image anomaly detection and localization network based on {S}imple{N}et and feature pyramid},
  author={Zhao, Yiming and Zhu, Fenghua and Mi, Yibo and Chen, Dewang and Xiong, Gang},
  booktitle={2024 IEEE 4th International Conference on Digital Twins and Parallel Intelligence (DTPI)},
  pages={417--422},
  year={2024},
  organization={IEEE}
}
}
\clearpage

\section*{Appendix}
Section \ref{app:dataset} reports details on image-based object distance estimation as well as distance-based statistics and information regarding the Argoverse 2 \cite{wilson2023argoverse} autonomous driving dataset.
Section \ref{app:eval} provides an additional evaluation of the proposed network and several baselines on the Argoverse dataset.
Section \ref{app:foveation} provides additional details regarding the hyperbolic foveation and its inverse, including computation times, existence of the inverse, and convergence guarantees. Lastly, Section \ref{app:architecture} provides additional details regarding the proposed network, Telescope, architecture while Section \ref{app:training_details} discusses training details.
\appendix

\section{Dataset Analysis}\label{app:dataset}
This section discusses the Argoverse 2 dataset \cite{wilson2023argoverse}.
This dataset contains objects in the near ($\leq 50$m), far ($50-150$m) and long ($150-250$m) ranges, however no ultra-long ($\geq 250$m) objects are present.
In Sec.~\ref{app_subsec:distance_estimation} we demonstrate how object ranges are computed using the object bounding box and camera parameters.
Next, we analyze the Argoverse dataset and the distribution of object ranges in Sec.~\ref{app_subsec:argoverse}.

\begin{figure}[b!]
    \centering
    \includegraphics[width=\linewidth]{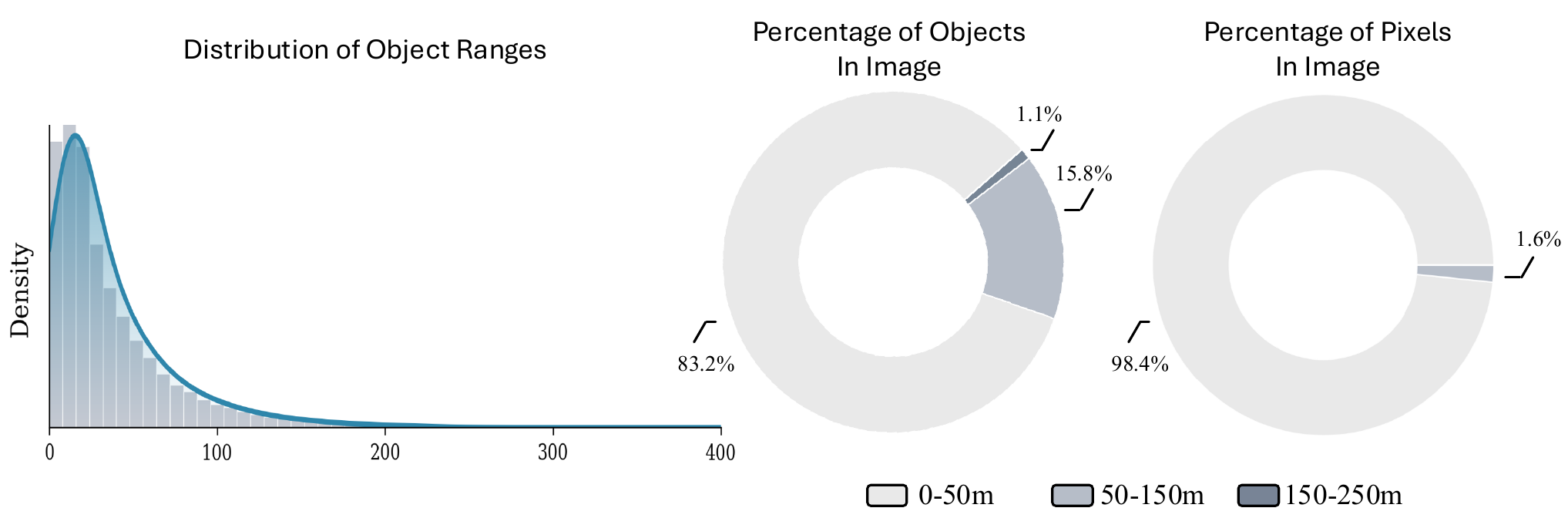}
    \caption{\textbf{Long-range Objects in Argoverse Driving.} Analysis of the Argoverse~\cite{wilson2023argoverse} dataset shows the distribution of object distances and the breakdown of the pixel-wise composition of objects at each distance. Multiple object ranges are represented in images, but nearby objects are disproportionately favored in terms of pixel area, with far (50-150m) and long (150-250m) range objects occupying only a small fraction of image pixels.}
    \label{fig:argoverse_dataset_analysis}
\end{figure}

\subsection{Distance Approximation from Bounding Boxes}
\label{app_subsec:distance_estimation}

Objects in the TruckDrive \cite{ghilotti2026truckdrive} dataset are annotated at distances extending up to $1\,\mathrm{km}$, where reliable LiDAR or radar measurements are often unavailable. 
Therefore, to estimate object distances, we follow \cite{ghilotti2026truckdrive} and approximate depth using the apparent size of the object in the image together with the camera intrinsics and an average class height prior.

Let $h_p$ denote the height of the detected bounding box in pixels, $f$ the camera focal length in pixels, and $H_c$ the average real-world height of the object class $c$ given in Table \ref{table:object_heights}. 
For the TruckDrive dataset, the focal length is $f=3304$.
Under the pinhole camera model, the object distance $d$ can be approximated as
\begin{equation}
d \approx \frac{f H_c}{h_p}.
\end{equation}
\vspace{-11pt}

\begin{table}[b] 
\centering
\scriptsize
\caption{\textbf{Average Class Heights}. Average heights across the class used to compute the approximate object distance given bounding box height and camera focal length.}
\label{table:object_heights}
\begin{tabularx}{\linewidth}{l*{6}{>{\centering\arraybackslash}X}}
\toprule
Class & Person & Bike & Car & Sign & Truck & Debris \\
\midrule \midrule
Average Height [m] & 0.70 & 0.71 & 1.89 & 1.26 & 2.90 & 0.41 \\

\bottomrule
\end{tabularx}
\end{table}

\subsection{Argoverse Dataset}\label{app_subsec:argoverse}
Similar to TruckDrive, an analysis of the Argoverse \cite{wilson2023argoverse} dataset is provided in Figure \ref{fig:argoverse_dataset_analysis}.
Notably, the Argoverse dataset only contains objects up to long range (e.g. $<250$m).
Nevertheless, we provide an ablation on Telescope and baseline performance to demonstrate the generalizability of the the proposed approach in Tables \ref{table:argoverse_distance_map} and \ref{table:argoverse_class_map}.
\section{Evaluation on Argoverse Dataset}\label{app:eval}

To demonstrate the generalizability of the proposed object detection model, Telescope, we perform further evalautions using the Argoverse dataset.
While this dataset does not contain ultra-long range objects, it does contain objects up to $<250$m.
To align with the evaluation in Section \ref{sec:exp}, object distances are estimated from bounding box height, camera intrinsics, and class-specific average object heights denoted in Table \ref{table:object_heights}. 
The camera focal length in pixels for the Argoverse 2 dataset is $1682$.

As in Section \ref{sec:exp}, an image resolution of $1024\times1024$ is used for all experiments.
We follow standard object detection protocols and report COCO-style mean average precision (mAP), together with distance-wise mAP computed over three distance bins
(\text{mAP}\textsubscript{0--50m}, \text{mAP}\textsubscript{50--150m}, and \text{mAP}\textsubscript{150--250m}).
We additionally report PASCAL-style mAP at IoU thresholds of 0.5 and 0.75.

Based on the findings of Table \ref{table:distance_map} and \ref{table:class_map}, we train and evaluate the \emph{three best-performing baselines}, Universenet \cite{shinya2021usb}, RVSA \cite{wang2022advancing}, and RFLA \cite{xu2022rfla}.
Baselines and Telescope are all fine-tuned for 5 epochs.

\begin{table}[t] 
\centering
\scriptsize
\caption{\textbf{Distance-Wise Object Detection Evaluation on Argoverse 2 Dataset}.
The proposed model using the SAM3 image encoder backbone, Deformable DETR detection head, de-noising training, and TeleScope re-sampling layer provides optimal performance across all distances.}
\label{table:argoverse_distance_map}
\resizebox{\linewidth}{!}{
\begin{tabular}{l|cccc|cc}
\toprule
\multirow{2}{*}{Method} & \multicolumn{4}{c|}{COCO} & \multicolumn{2}{c}{PASCAL} \\
 & mAP & mAP\textsubscript{0-50} & mAP\textsubscript{50-150} & mAP\textsubscript{150-250} & mAP\textsubscript{50} & mAP\textsubscript{75} \\
\midrule \midrule
UniverseNet \cite{shinya2021usb} & \cellcolor{tabsecond} 0.123 & \cellcolor{tabsecond} 0.156 & \cellcolor{tabthird} 0.042 & 0.016 & \cellcolor{tabsecond} 0.271 & \cellcolor{tabsecond} 0.100 \\
RVSA \cite{wang2022advancing} & \cellcolor{tabthird} 0.121 & \cellcolor{tabthird} 0.150 & \cellcolor{tabsecond} 0.052 & \cellcolor{tabsecond} 0.026 & \cellcolor{tabthird} 0.260 & \cellcolor{tabthird} 0.098 \\
RFLA \cite{xu2022rfla} & 0.106 & 0.131 & 0.036 & \cellcolor{tabthird} 0.023 & 0.250 & 0.070 \\
Telescope (Ours) & \cellcolor{tabfirst} 0.232 & \cellcolor{tabfirst} 0.268 & \cellcolor{tabfirst} 0.104 & \cellcolor{tabfirst} 0.036 &  \cellcolor{tabfirst} 0.502 & \cellcolor{tabfirst} 0.177 \\
\bottomrule
\end{tabular}
}
\end{table}
\begin{table}[t] 
\centering
\scriptsize
\caption{\textbf{Class-Wise Object Detection Evaluation on Argoverse 2 Dataset}. Leveraging foundation models as pre-trained image encoders provides a strong prior for object detection comparable to state-of-the-art specialized models. The proposed model using the SAM3 image encoder backbone, Deformable DETR detection head, de-noising training, and TeleScope re-sampling layer provides optimal performance across all classes. Shown are results for the 7 most common classes in Argoverse 2.}
\label{table:argoverse_class_map}

\setlength{\tabcolsep}{3pt}
\renewcommand{\arraystretch}{1.0}

\begin{tabularx}{\linewidth}{l|*{7}{>{\centering\arraybackslash}X}}
\toprule
\multirow{3}{*}{Method} & \multicolumn{7}{c}{COCO mAP} \\
 & Regular & Pedest. & Bollard & Const. & Stop & Bicycle & Wheeled \\
 & Vehicle &  &  & Barrel & Sign &  & Device
\\

\midrule \midrule

UniverseNet \cite{shinya2021usb} & \cellcolor{tabsecond} 0.396 & \cellcolor{tabsecond} 0.251 & \cellcolor{tabthird} 0.069 & \cellcolor{tabsecond} 0.353 & \cellcolor{tabsecond} 0.209 & \cellcolor{tabsecond} 0.151 & \cellcolor{tabsecond} 0.098 \\
RVSA \cite{wang2022advancing} & \cellcolor{tabthird} 0.381 & 0.183 & \cellcolor{tabsecond} 0.099 & 0.192 & 0.120 & \cellcolor{tabsecond} 0.151 & \cellcolor{tabthird} 0.091 \\
RFLA \cite{xu2022rfla} & 0.352 & \cellcolor{tabthird} 0.233 & 0.044 & \cellcolor{tabthird} 0.269 & \cellcolor{tabthird} 0.153 & \cellcolor{tabthird} 0.124 & 0.070 \\
Telescope (Ours) & \cellcolor{tabfirst} 0.562 & \cellcolor{tabfirst} 0.412 & \cellcolor{tabfirst} 0.154 & \cellcolor{tabfirst} 0.417 & \cellcolor{tabfirst} 0.316 & \cellcolor{tabfirst} 0.355 & \cellcolor{tabfirst} 0.259 \\

\bottomrule
\end{tabularx}
\end{table}

\begin{figure*}[h]
    \centering
    \includegraphics[width=\linewidth]{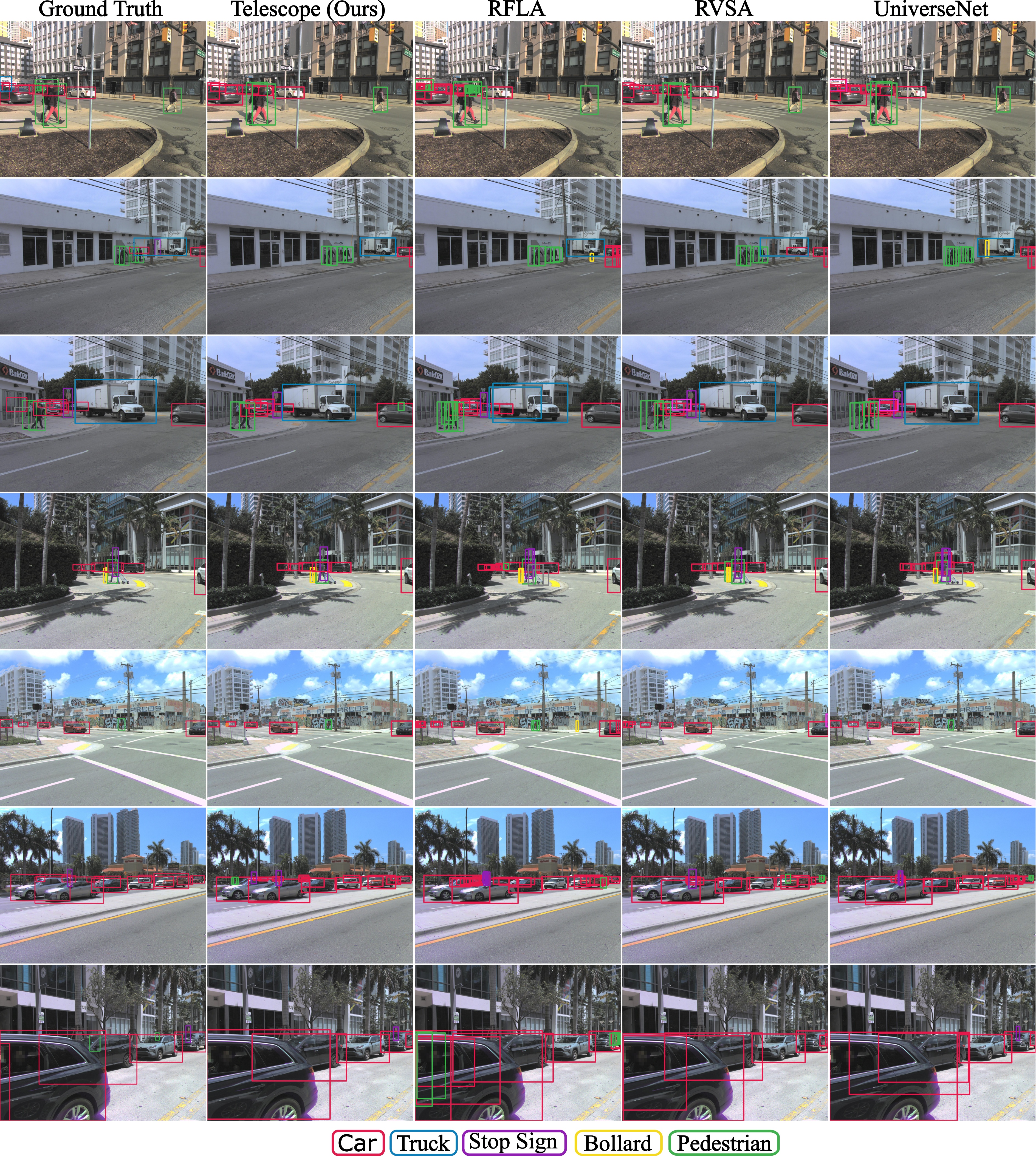}
    \caption{\textbf{Additional Qualitative Comparison on Argoverse Dataset.} Qualitative comparison between the proposed method, Telescope, and state-of-the-art baselines specialized for small object detection. Ground truth annotations are shown on the left. Notably, there are many target boxes which represent occluded objects (rows 1, 2, 3, 6, and 7). All methods are fine-tuned on the Argoverse \cite{wilson2023argoverse} dataset.}
    \label{fig:additional_qual_argo}
\end{figure*}

\begin{figure*}[h]
    \centering
    \includegraphics[width=\linewidth]{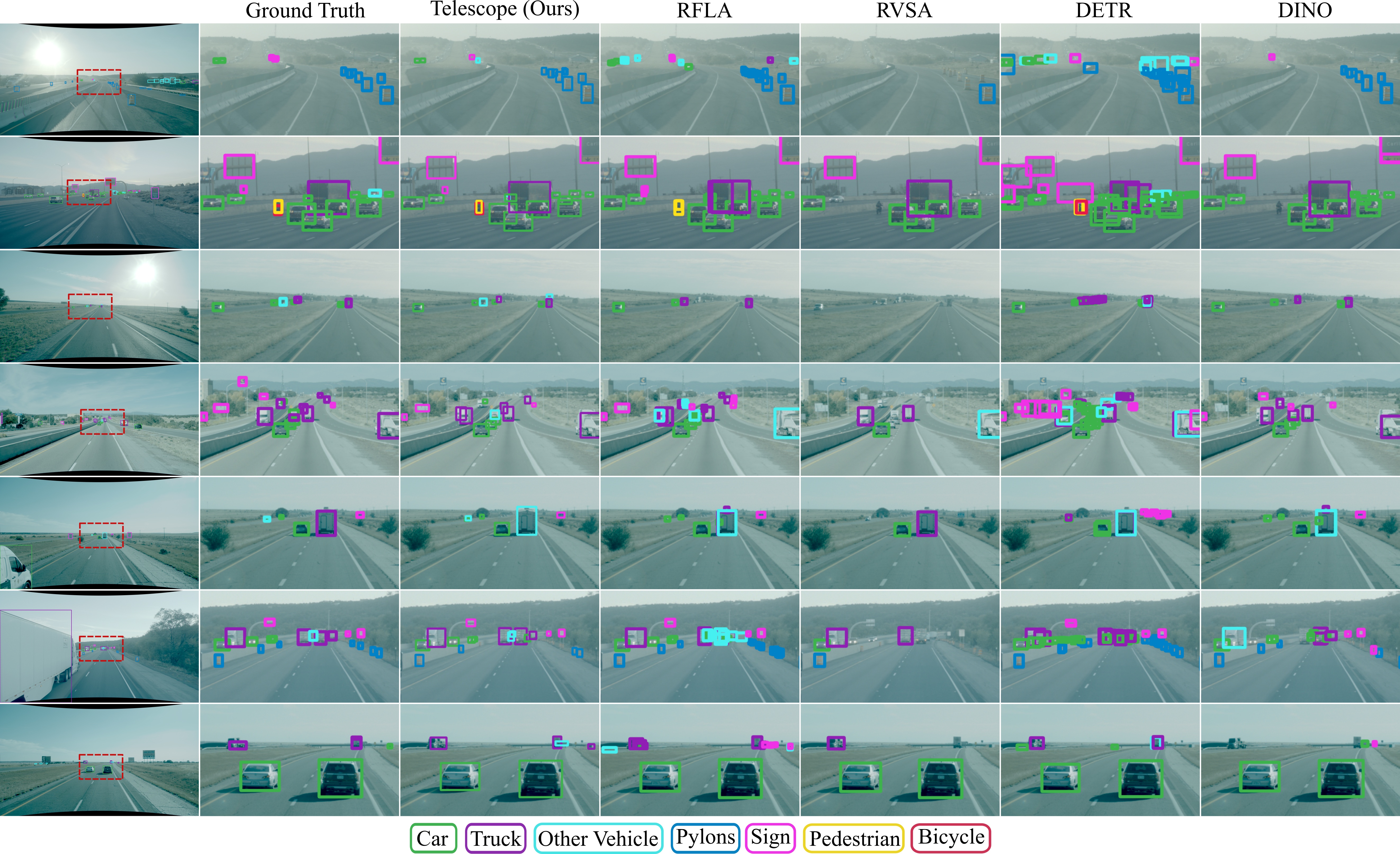}
    \caption{\textbf{Additional Qualitative Comparison on TruckDrive Dataset.} Qualitative comparison between the proposed method, Telescope, and state-of-the-art baselines. Both RVSA \cite{wang2022advancing} and RFLA \cite{xu2022rfla} are specialized for small object detection while DETR \cite{carion2020detr} and DINO \cite{zhang2022dino} are strong general object detectors, but perform worse in long an ultra-long range object detection. Ground truth annotations are shown on the left. Zoomed-in views corresponding to the red rectangles are provided to highlight detections at long and ultra-long range, where some objects reach up to $1$km. All baselines are fine-tuned on the TruckDrive \cite{ghilotti2026truckdrive} dataset.}
    \label{fig:additional_qual_truck}
\end{figure*}

As reported in Tables \ref{table:argoverse_distance_map} and \ref{table:argoverse_class_map}, we find that the proposed model, Telescope, outperforms all baselines across all distance ranges and classes.
Evaluations are computed in the same manner as in Tables \ref{table:distance_map} and \ref{table:class_map} with TruckDrive.
This reflects the findings from Section \ref{sec:exp}, confirming the efficacy and generalizability of Telescope across both the TruckDrive \cite{ghilotti2026truckdrive} and Agoverse \cite{wilson2023argoverse} datasets.
Qualitative results for the Argoverse dataset are presented in Figure \ref{fig:additional_qual_argo}, along with additional qualitative results from the TruckDrive dataset in Figure \ref{fig:additional_qual_truck}.

\section{Hyperbolic Foveation Computation}\label{app:foveation}

We first prove the existence of the inverse of the hyperbolic foveated transform.
We then show that this inverse is approximated using the Newton-Raphson algorithm with guarantees on convergence.

\paragraph{Theorem 1 (Existence of the Inverse).}
Assume $\alpha, p, R>0$.  
Then $\Phi$ is a diffeomorphism on $\mathbb{R}^2$.

\paragraph{Proof.}
For $r>0$, the map in~\eqref{eq:foveated_map} is a smooth radial deformation centered at $o$ with strictly positive radial derivative
$
\partial_r \| \Phi(x)-o \| > 0
$
since both the hyperbolic contraction $\tanh(\alpha r)$ and the interpolation weight $w(r)$ are monotone in $r$.  

The Jacobian of $\Phi$ is everywhere non-singular, implying local invertibility.  
Global injectivity follows from strict radial monotonicity, and surjectivity follows from $\Phi(x)=x$ for $r\ge R$.  
Hence $\Phi$ is a diffeomorphism. \hfill$\square$

\paragraph{Theorem 2 (Convergence of the inverse approximation).}
Let $y=\Phi(x^\star)$.  
If the Jacobian $J_\Phi(x^\star)$ is non-singular, then the Newton--Raphson iteration
$
x^{(k+1)} = x^{(k)} - J_\Phi(x^{(k)})^{-1}(\Phi(x^{(k)})-y)
$
converges locally and quadratically to $x^\star$.

\paragraph{Proof.}
Since $\Phi$ is continuously differentiable and $J_\Phi(x^\star)$ is invertible by Theorem~1, the standard Newton--Raphson convergence theorem applies, yielding local quadratic convergence. \hfill$\square$

\paragraph{Transform Runtime}
We also analyze the computation time for the forward and backwards transform.
We randomly initialize 100 boxes, use a batch size of 4, and run 50 transformations.
The Euclidean to Riemannian (forward) transformation takes $1.86 \pm 0.08$ms. For the Riemannian to Euclidean (backwards) transformation ir takes approximately $8$ Newton-Raphson iterations to achieve an error tolerance of $<1e-06$, which combined take $16.6 \pm 2.73$ms.

\section{Network Architecture Details}\label{app:architecture}
\begin{table}[t]
\centering
	\caption{
    Network Details of SAM3 + DINO 2-Stage Model.
    }
    \resizebox{0.98\linewidth}{!}{
    \resizebox{\linewidth}{!}{
\begin{tabular}{c|c|c|c}
        \textbf{Component} & \textbf{Sub-Component} & \textbf{Layer} & \textbf{Parameters} \\
        \hline
        \hline
         \multirow{4}{*}{SAM3 Backbone} & \multirow{2}{*}{ViT (ViTDet)} & Patch Embed & patch: 14, dim: 1024\\
          &  & Transformer & depth: 32, heads: 16, mlp: 4.625$\times$\\
        \cline{2-4}
          & \multirow{2}{*}{FPN Neck} & ConvTranspose2d & $\times$2 upsample per scale\\
          &  & Conv2d & (1024, 256), k=1; (256, 256), k=3\\
         \hline

         \multirow{3}{*}{Foveation Estimation} & \multirow{2}{*}{Head} & AdaptiveAvgPool2d & output: 1$\times$1\\
          &  & MLP & (256, 128, 4), ReLU, Sigmoid/Softplus\\
        \cline{2-4}
          & Foveation Embed & MLP & (4, 64, 256), ReLU\\
         \hline

         \multirow{6}{*}{DINO Transformer} & Input Proj & Conv2d + GN & (256, 256), k=1, GN(32)\\
        \cline{2-4}
          & \multirow{2}{*}{Encoder $\times 6$} & MSDeformAttn & heads: 8, levels: 3, points: 4\\
          &  & FFN & (256, 2048, 256), ReLU\\
        \cline{2-4}
          & \multirow{2}{*}{Decoder $\times 6$} & MSDeformAttn & heads: 8, levels: 3, points: 4\\
          &  & FFN & (256, 2048, 256), ReLU\\
        \cline{2-4}
          & Query Selection & Top-k & k=300 proposals from encoder\\
         \hline

         \multirow{3}{*}{Detection Head} & Class Embed & Linear & (256, $C$)\\
          & Bbox Embed & MLP & (256, 256, 4), 3 layers\\
          & Label Embed (DN) & Embedding & ($C$+1, 256)\\
         \hline
    \end{tabular}
}
    \label{table:network_sam3_dino}}
\end{table}

We leverage the SAM3 image encoder for this work.
This encoder is derived from the Perception Encoder \cite{bolya2025perception} and uses a ViT (Vision Transformer) with 32 layers, an embedding dimension of 1024, and 14×14 patches.
Windowed local self-attention is used with global self-attention every 7th layer, alongside 2D RoPE and a SimpleFPN neck \cite{zhao2024simple} that produces 256-dimensional feature maps at $4\times$, $2\times$, and $1\times$ resolutions.
The ViT backbone and FPN are frozen for all experiments. 

The proposed model illustrated in Figure 2 of the main manuscript, Telescope, consists of two stages.
In the first stage, the input image is down-sampled to $512 \times 512$ and passed through the SAM3 image encoder.
The $1\times$ feature resolution outputs are then flattened and run through a 3-layer MLP which estimates the four foveation parameters.
These parameters are the foveation center, $[c_x, c_y]$, and the foveation radius, $R_x$ and $R_y$.
For simplicity, the maximum radius is used (i.e. $\max(R_x, R_y)$.

In the second stage, the hyperbolic foveated transform is applied to the original resolution image.
The transformed image is then down-sampled to $1024\times 1024$ and passed through the SAM3 image encoder.
All three feature resolutions are then used as inputs to a Deformable DETR \cite{zhu2020deformable} detection head consisting of a $256$-d deformable encoder and decoder with 4 sampling points per level. 
Two-stage refinement is used from encoder proposals.
A 3-layer MLP head then estimates the Riemannian bounding box parameters as discussed in Section \ref{sec:method}. 
See Table \ref{table:network_sam3_dino} for details on the network parameters.
\clearpage
\onecolumn
\twocolumn
\section{Training Details}\label{app:training_details}
In our training, we use the denoising proposed in \cite{zhang2022dino}, where ground truth bounding box parameters are first noised, concatenated with the object queries, and then de-noised by the detection head.
This helps stabilize the matching process during learning and provides an early training signal to the detection head for bounding box localization.
For Telescope, the Euclidean ground truth boxes are first noised, then projected to the Riemannian space via \eqref{eq:foveated_map} and appended to the queries.
This ensures that these boxes remain in the same space as the network predictions.

For all Telescope experiments, we used a learning rate of $1e-04$, a lambda learning rate schedule with $1$ warm-up epoch, a batch size of $4$, $300$ decoder queries, and trained across $2$ A100 GPUs.
For baselines, the default training parameters specified in the publicly available repos were used.

\end{document}